\documentclass[journal]{arxiv}
\let\labelindent\relax

\usepackage[colorlinks,urlcolor=blue,linkcolor=blue,citecolor=blue]{hyperref}

\usepackage{color,array}

\usepackage{graphicx}

\usepackage{soul}
\usepackage{url}
\usepackage[utf8]{inputenc}
\usepackage[small]{caption}
\usepackage{amsthm}
\usepackage{booktabs}
\usepackage[switch]{lineno}

\usepackage{graphicx}
\usepackage{textcomp}
\usepackage[square,sort,comma,numbers]{natbib}
\usepackage{xcolor}
\def\BibTeX{{\rm B\kern-.05em{\sc i\kern-.025em b}\kern-.08em
    T\kern-.1667em\lower.7ex\hbox{E}\kern-.125emX}}
\usepackage{url}
\usepackage{fancyhdr,amsmath,graphicx} 
\usepackage{amsfonts}       %
\usepackage{colortbl}
\usepackage{multirow,makecell}
\usepackage{algpseudocode}
\usepackage{graphics}
\usepackage{epsfig}
\usepackage{xfrac}
\usepackage[ruled,vlined,linesnumbered]{algorithm2e}
\usepackage{color}
\usepackage{float}
\usepackage{subfig}
\usepackage{xspace}
\usepackage{soul}
\usepackage{enumitem}

\newcommand{{\fgu}}{{{\textsf{\small{FGU}}}\xspace}}
\usepackage{lipsum,booktabs}
\definecolor{myblue}{rgb}{0,0.05,0.4}
\usepackage[capitalize,noabbrev,nameinlink]{cleveref}

\newcommand{\cora}{{\small\texttt{Cora}}\xspace}
\newcommand{\citeseer}{{\small\texttt{Citeseer}}\xspace}

\newcommand{{\grapheraser}}{{{\textsf{\small{GraphEraser}}}\xspace}}
\newcommand{{\retrain}}{{{\textsf{Retrain}}\xspace}}
\newcommand{{\ges}}{{{\textsf{\small{GEraser}}}\xspace}}
\newcommand{{\gedit}}{{{\textsf{\small{GEditor}}}\xspace}}
\newcommand{{\gif}}{{{\textsf{\small{GIF}}}\xspace}}
\newcommand{{\gd}}{{{\textsf{\small{GDelete}}}\xspace}}
\newcommand{{\sgc}}{{{\textsf{\small{SGC}}}\xspace}}
\newcommand{\ms}[2]{{#1\footnotesize{$\pm$#2}}}

\newcommand{\german}{\texttt{German}\xspace}
\newcommand{\bail}{\texttt{Bail}\xspace}
\newcommand{\credit}{\texttt{Credit}\xspace}
\newcommand{\pokecz}{\texttt{Pokec-z}\xspace}
\newcommand{\pokecn}{\texttt{Pokec-n}\xspace}
\newcommand{\nba}{\texttt{NBA}\xspace}

\newcommand{\graph}{\ensuremath{\mathcal{G}}\xspace}
\newcommand{\partition}{\ensuremath{\mathcal{P}}\xspace}
\newcommand{\model}{\ensuremath{\mathcal{M}}\xspace}
\newcommand{\utilityloss}{\ensuremath{\mathcal{U}}\xspace}
\newcommand{\fairloss}
{\ensuremath{\mathcal{F}}\xspace}
\newcommand{\totalloss}
{\ensuremath{\mathcal{L}}\xspace}
\newcommand{\nodeset}{\ensuremath{\mathcal{V}}\xspace}
\newcommand{\edgeset}{\ensuremath{E}\xspace}
\newcommand{\feat}{\ensuremath{X}\xspace}
\newcommand{\adj}{\ensuremath{A}\xspace}
\newcommand{\senattri}{\ensuremath{S}\xspace}
\newcommand{\glabel}{\ensuremath{Y}\xspace}

\newcommand{\neigh}[1]{\ensuremath{\mathcal{N}_{#1}}\xspace}

\usepackage{tikz}
\newcommand{\ballnumber}[1]{\tikz[baseline=(myanchor.base)] \node[circle,fill=.,inner sep=1pt] (myanchor) {\color{-.}\bfseries\scriptsize #1};}

\AtBeginDocument{%
  \providecommand\BibTeX{{%
    \normalfont B\kern-0.5em{\scshape i\kern-0.25em b}\kern-0.8em\TeX}}}

\definecolor{mygray}{gray}{0.9}
\definecolor{green}{rgb}{0.1,0.1,0.1}

\definecolor{LightCyan}{rgb}{0.88,1,1}
\definecolor{Gray}{gray}{0.85}
\newcolumntype{a}{>{\columncolor{Gray}}c}
\definecolor{myorange}{RGB}{255,200,150}
\sethlcolor{LightCyan}

\begin{document}

\title{Enabling Group Fairness in Graph Unlearning via Bi-level Debiasing}

\author{Yezi Liu, Prathyush Poduval, Wenjun Huang, Yang Ni, Hanning Chen, and Mohsen Imani%
\thanks{Yezi Liu is with the Department of Electrical Engineering and Computer Science, University of California, Irvine, CA 92697 USA (e-mail: yezil3@uci.edu).}%
\thanks{Prathyush Poduval, Wenjun Huang, Yang Ni, Hanning Chen, and Mohsen Imani are with the Donald Bren School of Information and Computer Sciences, University of California, Irvine, CA 92697 USA (e-mail: ppoduval@uci.edu; wenjunh3@uci.edu; yni3@uci.edu; hanningc@uci.edu; m.imani@uci.edu).}%
\thanks{Corresponding author: Mohsen Imani (e-mail: m.imani@uci.edu).}
}

\maketitle

\begin{abstract}
Graph unlearning is a crucial approach for protecting user privacy by erasing the influence of user data on trained graph models. Recent developments in graph unlearning methods have primarily focused on maintaining model prediction performance while removing user information. However, we have observed that when user information is deleted from the model, the prediction distribution across different sensitive groups often changes. Furthermore, graph models are shown to be prone to amplifying biases, making the study of fairness in graph unlearning particularly important. This raises the question: \textit{Does graph unlearning actually introduce bias?} Our findings indicate that the predictions of post-unlearning models become highly correlated with sensitive attributes, confirming the introduction of bias in the graph unlearning process. To address this issue, we propose a fair graph unlearning method, {\fgu}. To guarantee privacy, {\fgu} trains shard models on partitioned subgraphs, unlearns the requested data from the corresponding subgraphs, and retrains the shard models on the modified subgraphs. To ensure fairness, {\fgu} employs a bi-level debiasing process: it first enables shard-level fairness by incorporating a fairness regularizer in the shard model retraining, and then achieves global-level fairness by aligning all shard models to minimize global disparity. Our experiments demonstrate that {\fgu} achieves superior fairness while maintaining privacy and accuracy. Additionally, {\fgu} is robust to diverse unlearning requests, ensuring fairness and utility performance across various data distributions.
\end{abstract}

\begin{IEEEkeywords}
Fairness, graph neural networks, machine unlearning
\end{IEEEkeywords}

\section{Introduction} 
\IEEEPARstart{W}{ith} the increasing user demands for data privacy~\citep{curzon2021privacy,huang2022overview}, regulations such as ``the right to be forgotten (RTBF)"~\citep{rosen2011right} have been introduced by governments all over the world to guarantee the requested data is removed from machine learning models~\citep{GDPR,CCPA,scantamburlo2024artificial}. Meanwhile, graph neural networks are particularly sensitive to privacy concerns, as they are widely used in social network analysis~\citep{hamilton2017inductive,gong2025scalable}, recommendation systems~\citep{ying2018graph}, drug discovery~\citep{bongini2021molecular}, and epidemiology~\citep{liu2024review}. To comply with these regulations for graph-based applications, many graph neural network unlearning methods (short for graph unlearning) have been proposed~\citep{chen2022graph,cheng2023gnndelete,cong2022grapheditor,klicpera2019combining}, which involve deleting graph data, such as nodes, edges, node attributes, and edge attributes.

\begin{figure}[t]
   {\includegraphics[height=1.4in]{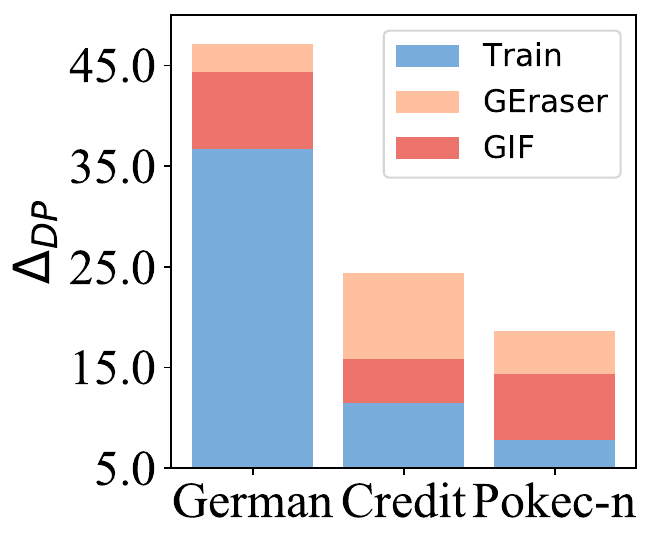}\label{intro:dp}}
    \hspace{-8pt}
    {\includegraphics[height=1.4in]{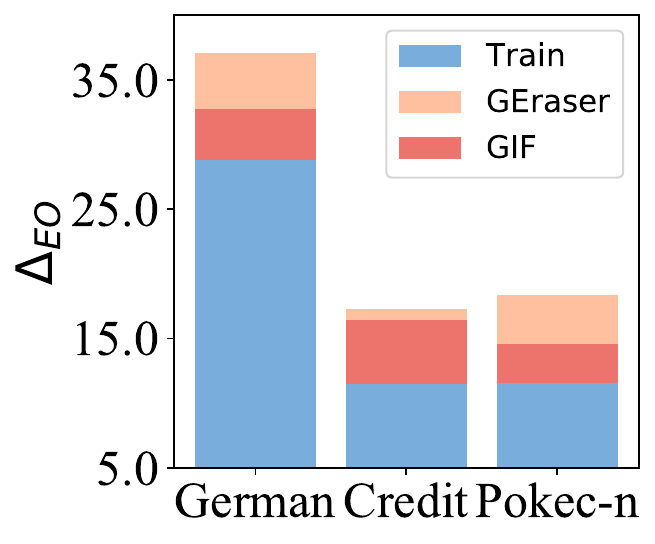}\label{intro:eo}}
    \caption{Fairness performance of a two-layer GCNs before (left) and after (right) applying {\ges} and {\gif}. $10\%$ of the nodes and $10\%$ of the edges were randomly sampled from the unprivileged sensitive group for unlearning.}
    \vspace{-1em}
    \label{fig:intro}
\end{figure}

Most previous efforts on graph unlearning studies have focused on the performance drop problem and progress to maintain good prediction accuracy while removing graph information. Whether the graph unlearning process leads to other problems, for example, the fairness issue, is less explored~\citep{mehrabi2021survey,dwork2012fairness}.
To address this gap, we conducted an empirical investigation to explore the relationship between graph unlearning and fairness. 
In~\cref{fig:intro}, we observed that two-layer graph convolution networks obtained a worse fairness performance on three datasets after unlearning $10\%$ of nodes and $10\%$ of edges. The popular unlearning methods {\ges} (orange) and {\gif} (red) resulted in an increase in bias than the original training (blue), as evidenced by the higher values of $\Delta_{DP}$ and $\Delta_{EO}$. Discrimination can lead to significant social and ethical issues~\citep{liu2025white}, limiting their applicability in real-world applications such as healthcare~\citep{ahmad2020fairness,chen2018my}, job recruitment~\citep{mehrabi2021survey}, credit scoring~\citep{kozodoi2022fairness}, and criminal justice~\citep{berk2021fairness}. This raises a critical question that anchors our study:
\begin{center}
\textbf{\textit{Can graph unlearning ensure fairness while \\ preserving utility?}}
\end{center}

Debiasing in graph unlearning is an open and challenging problem. Graph unlearning can be divided into two categories: exact and approximate graph unlearning. The \textit{challenges} arise from the specific design and requirements of these methods. 1) Exact graph unlearning involves splitting a graph into smaller subgraphs, training a model on each of these smaller parts (also called shards), and then combining the results from each sub-model~\citep{cong2022grapheditor,chen2022graph}. When a deletion request is made, these methods remove the data from the relevant shard and retrain the shard models. To introduce fairness, debiasing techniques can be applied during the retraining of each shard model. However, this approach only ensures fairness within individual shard models. Bias can still be introduced during the aggregation step, even if each shard model is fair. 2) Approximate graph unlearning adjusts the trained model when a data removal request is made, aiming to approximate the results of retraining the model on the remaining data.~\citep{wu2023gif,cheng2023gnndelete}. These methods are difficult to optimize, and adding fairness considerations makes the optimization process even more challenging.

\begin{figure*}[]
\centering
\includegraphics[width=1.0\linewidth]{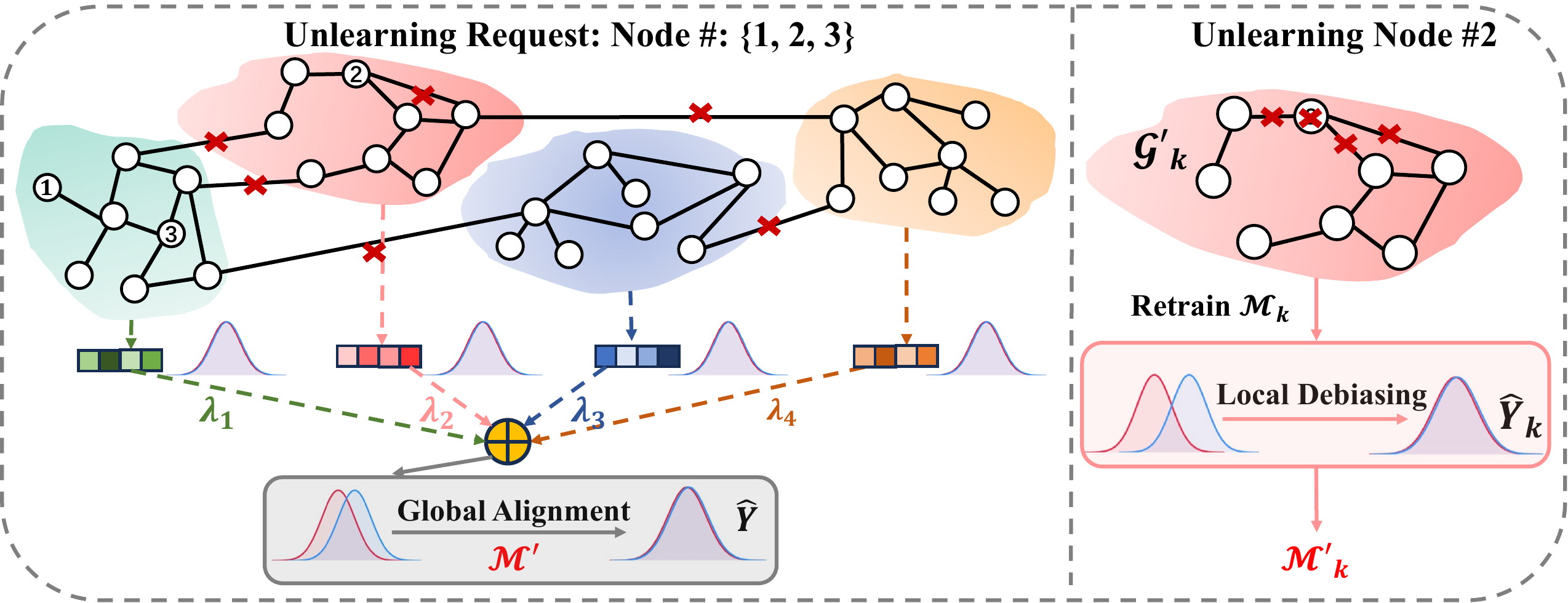}
\caption{An illustration of {\fgu}: \textit{\textbf{shard debiasing}} achieves equitable predictions within each shard model, and \textit{\textbf{global alignment}} reduces disparities among sensitive groups across shard models.} \label{fig:framework}
\vspace{-1em}
\end{figure*}

To tackle these challenges, we propose a fairness-aware graph unlearning approach named {\fgu}. Our approach aims to ensure fairness and privacy while maintaining the utility of the models after unlearning. {\fgu} adopts a bi-level debiasing strategy. First, it incorporates a shared fairness regularizer into the shard retraining process. Then, {\fgu} aligns the shard models to reduce global disparities among sensitive groups. We outline \textbf{our contributions} as follows:
\begin{itemize}[leftmargin=0.35cm, itemindent=.0cm, itemsep=0.0cm, topsep=0.0cm]
\item We pioneer a study of fairness issues in existing graph unlearning methods, based on the following observations: \textit{(i)} Both exact and approximate graph unlearning methods introduce bias to the post-unlearning model. \textit{(ii)} Exact graph unlearning methods tend to introduce more bias compared to approximate graph unlearning methods. \textit{(iii)} Higher unlearning ratios introduce more bias, and unlearning data from the unprivileged group induces more bias than unlearning data from the privileged group.
\item We formally define the \textit{fairness problem} on graph unlearning. To the best of our knowledge, the proposed {\fgu}, is the first graph unlearning framework to simultaneously preserve the privacy of requested data for a trained model and ensure the fairness of the post-unlearning model. {\fgu} achieves fairness through bi-level debiasing, eliminating the need to retrain the entire remaining dataset. 
\item Comprehensive experiments across six datasets demonstrate that {\fgu} effectively protects user privacy, maintains model performance, and ensures fairness. Moreover, {\fgu} demonstrates stable performance across diverse unlearning requests, which include various data distributions from the original dataset.
\end{itemize}
\vspace{-0.5em}
\section{Preliminary}
\noindent\textbf{Notations of GNNs.}
Let $\graph = (\nodeset, \edgeset, \feat,\senattri)$ denote a graph, where $\nodeset = \{v_1, v_2, \cdots v_n\}$ represents a set of nodes, and $\edgeset = \{(v_i, v_j)\}$ denotes a set of edges, indicating pairwise associations between nodes. $\feat \in \mathbb{R}^{N \times D}$ is the feature matrix with $D$ dimension, and ${A} \in\{0,1\}^{N \times N}$ is the adjacency matrix, indicates the presence or absence of edges between nodes. $\senattri\in\{0,1\}^N$ is a vector containing sensitive attributes (e.g., gender or race) of nodes that a graph model should not capture to make decisions. $\glabel = \{{\glabel}_1,{\glabel}_2,..,{\glabel}_n\}$ consists of the labels associated with each node in the graph. The message passing process of GNN can be found in~\cref{appendix:gnn}.

\noindent\textbf{Graph Unlearning Setups.}
Consider a trained GNN model $\model$ parameterized on weights $\theta$ that generate prediction $\hat{\glabel}=\{\hat{\glabel}_1,\hat{\glabel}_2,..,\hat{\glabel}_n\}$. To satisfy the RTBF, the users request to remove their personal data, which can be categorized into unlearned node set $\nodeset_u$, unlearned edge set $\edgeset_u$, and unlearned features $\feat_u$ from the users. A graph unlearning framework will output with a post-unlearning model $\model'$, which generates a prediction $\hat{\glabel}'$. In this paper, we focus on the \emph{node classification task}, whose goal is to use a GNN to predict the label of a node $i \in \nodeset$ given the node's features $\feat_i$ and information of its neighbors. Detailed related works on graph unlearning are elaborated in \cref{appendix:graph_unlearning}.

\noindent\textbf{Group Fairness Definitions.}\label{appendix:group_fair}
Demographic parity (DP) and equal opportunity (EO) are two fundamental concepts in group fairness~\citep{liu2023fairgraph}. Demographic parity~\citep{dwork2012fairness} requires that a classifier assigns positive outcomes at the same rate across different groups. Equal opportunity~\citep{hardt2016equality} focuses on the performance of a binary predictor $\hat{\glabel}$ in relation to the ground truth labels $\glabel$ and sensitive attributes $\adj$. These fairness concepts are formulated as:
\begin{equation}
\begin{array}{c}
\Delta_{EO}\!=\!\vert P(\hat{\glabel}\!=\!1 \!\!\mid\!\! A\!=\!0, Y\!=\!1)\!-\!P(\hat{\glabel}\!=\!1 \!\!\mid\!\! A\!=\!1, Y\!=\!1) \vert, \\
\Delta_{DP}=\vert P(\hat{\glabel}=1 \mid A=0)-P(\hat{\glabel}=1 \mid A=1) \vert,~\label{eq:fairness}
\end{array}
\end{equation}
where lower values of \(\Delta_{DP}\) and \(\Delta_{EO}\) indicate better fairness.

\noindent\textbf{The Fair Graph Unlearning Problem.} The \textit{fair graph unlearning} problem aims to update the model $\model$ and obtain a fair model $\model'$ after removing either nodes $\nodeset_u$, node feature vectors $\feat_u$, and/or edges $\edgeset_u$. The fair model $\model'$ should have the comparable utility, i.e., prediction performance, with the GNN trained on the original graph $\graph$; and its prediction $\hat{\glabel}\in\{0,1\}^N$ should satisfy the group fairness in \cref{appendix:group_fair}. 
\section{FGU: Fairness-aware Graph Unlearning}
In this section, we present the proposed {\fgu} framework. We begin by introducing the shard training process, including shard importance learning. Next, we explain the graph unlearning process, which involves node and/or edge unlearning. Then, we describe the debiasing mechanism of {\fgu}. Finally, we detail the optimization strategy of the algorithm. \cref{fig:framework} shows an overview of {\fgu}.

\subsection{Shard Training and Aggregation}\label{method:graph_partition_training} 
\noindent\textbf{Shard Training.} 
Shard training and retraining an efficient method that eliminates the need to retrain the entire model, which involves optimizing a large number of parameters. It has been used in unlearning studies, for example, in machine unlearning~\citep{liu2021federaser,wu2022federated} and in graph unlearning~\citep{chen2022graph,wang2023inductive}. This approach is particularly advantageous in the context of graph neural networks, which are extremely difficult to train. In graph unlearning, where the goal is to achieve fairness and privacy without sacrificing efficiency, shard training significantly enhances the algorithm's performance, making it more suitable for practical applications.

We implement shard training on $\graph$, a method initially proposed in a machine unlearning framework SISA~\citep{bourtoule2021machine} and widely used in graph unlearning studies~\citep{chen2022graph,wang2023inductive,dukler2023safe}. This process begins by partitioning $\graph$ into $k$ shards using a partitioning algorithm $\partition$. Following the approach in previous work~\citep{chen2022graph}, {\fgu} adopts the \textit{balanced graph partition algorithm}, which partitions the graph $\graph$ into a set of shard graphs, denoted as $\graph_p=\partition(\graph) = \{\graph_1, \graph_2, \cdots, \graph_k\}$, where each $\graph_k$ is a subgraph of $\graph$, referred to as the shard graph for shard $k$. Subsequently, a shard model $\model_k$ is trained on each shard graph $\graph_k$. The objective of the shard training process is defined as:
\begin{align}
&\underset{\lambda,\theta}{\min} \;  \underset{i\in\nodeset_0}{\mathrm{E}}
\left[\utilityloss_k(\theta_k) \left( \sum_{k=0}^{K} \lambda_k \model_{k}(\feat_i, \neigh{i}), \glabel_i \right) \right],\label{eq:shard_training}\\
&\text{with}  \ \  \utilityloss_k(\theta_k) = \sum_{i\in \nodeset_k} -[\glabel_i \log(\hat{\glabel}_i) + (1-\glabel_i)\log (1-\hat{\glabel}_i)],\label{eq:utility_loss}
\end{align}
where $\utilityloss_k(\theta_k)$ denotes the utility loss of shard $k$ and we adopt the cross-entropy loss in {\fgu}. $\lambda = \{\lambda_1, \lambda_2, \cdots \lambda_K\}$ indicates the importance scores of the shard models, satisfying $\lambda_k \geq 0$, and $\sum{\lambda_k} = 1$. $\nodeset_0$ is randomly sampled from $\nodeset$. $\neigh{i}$ is the neighborhood of node $i$ and $\gamma$ is the regularization parameter. The prediction $\hat{\glabel}_i$ is obtained by GNN model $M$ (\cref{appendix:gnn}) on node $i$.

\noindent\textbf{Aggregation and Inference.} The most straightforward aggregation strategy, primarily used in~\citep{bourtoule2021machine}, is majority voting: If the posteriors of the shard models show high confidence in multiple classes rather than a single class, the majority voting can lose information about the runner-up classes, leading to poor model utility. Another method is mean aggregation, which involves averaging the posteriors of a node from all shard models. This posterior is a vector indicating the probability of the node belonging to each class, with the sum of all values in the vector equal to 1.

In the learning-based aggregation used in {\fgu}, the process works differently. During inference, to predict the label of node $i$, the model $\model$ sends the data (features $\feat_i$ of node $i$ and the graph structure) to all shard models. Each shard model then makes a prediction, resulting in a posterior vector for node $i$. The final prediction for node $i$ is determined by taking the weighted sum of all posterior vectors from the shard models, using the learned importance $\lambda$.
\subsection{Standard Graph Unlearning}
\noindent\textbf{Node and Edge Unlearning.} The graph unlearning for the node classification task is specifically the node and edge unlearning. When the deletion request comes, the \textit{first step} of standard graph unlearning is to remove the requested edges and nodes from the corresponding shard graph and obtain the altered shard data $\graph=\{\graph'_1,\cdot,\graph'_K\}$. 
Specifically, for \textit{node unlearning} only, assume the requested node set is 
$\nodeset_u$, the updated graph is $\graph'=  (\nodeset\setminus\nodeset_u, \edgeset', \feat\setminus\feat_u, \senattri\setminus{S}_{u} )$, where $\edgeset'$ contains edges connected with all nodes in $\nodeset_u$.
For \textit{edge unlearning} only, assume the requested edge set is $\edgeset_u$, the updated graph is $\graph'= (\nodeset, \edgeset \setminus \edgeset_u, \feat, \senattri)$. For \textit{feature unlearning} only, given the unlearned feature set $\feat_u$, the updated graph is $\graph'= (\nodeset, \edgeset, \feat \setminus \feat_u, \senattri)$. In this paper, we consider a mixed unlearning for nodes and edges. 

\noindent\textbf{Standard Shard Retraining.} The \textit{second step} of standard graph unlearning involves shard retraining, where we replace the graph in \cref{eq:shard_training} with the updated graph $\graph'$. So far, we have described the standard graph unlearning process, specifically the exact graph unlearning without considering fairness, which outputs the retrained model $\model'$ for downstream tasks. The shard retraining property actually facilitates the incorporation of debiasing modules in graph unlearning methods. In the following subsection, we will elaborate on how to enable group fairness in the graph unlearning process.
\subsection{Enabling Fairness in Graph Unlearning}
The fairness definitions in \cref{appendix:group_fair} are straightforward to apply in centralized model training to evaluate model performance. However, in graph unlearning, the data in shards often have non-IID distributions, necessitating different levels of fairness consideration: \textit{global fairness} and \textit{shard fairness}. \textit{Global fairness} evaluates the fairness of a model across the entire graph $\graph = \cup_k \graph_k$, spanning all $K$ shards in graph unlearning. Our ultimate goal in fair graph unlearning is to train a model that is non-discriminatory to any group within the full graph. \textit{Shard fairness}, on the other hand, assesses the fairness of individual shard models $\graph_k$. This shard fairness is fundamental to achieving global fairness, as the final model is an aggregation of these shard models.

\noindent\textbf{Shard Retraining and Debiasing.}
For achieving \textit{shard fairness}, after deleting the requested data from the corresponding subgraph, the shard models are retrained on the remaining data. To ensure fairness, in each retraining epoch, {\fgu} incorporates the shard fair loss into the shard retraining. We consider $\Delta_{DP}$, defined in~\cref{eq:fairness}, as the regularizer and shard fair loss is based on it for each shard model. The shard fair loss is then used to update the shard model. The shard fair loss is calculated as follows:
\begin{align}
      &\fairloss_{k} = \vert P(\hat{\glabel}'_{k}=1|\senattri'_{k}=0) - P(\hat{\glabel}'_{k}=1|\senattri'_{k}=1)\vert,  \label{eq:local_dp} 
\end{align}
{\fgu} optimizes the unlearned shard model $\model'_k$ on $\totalloss_{k}$ to obtain the unbiased shard models.

\noindent\textbf{Model Aggregation and Global Alignment.} We aggregate the shard model taking the weighted sum of the shard model weights: ${\tilde{\theta}} = 
\sum_{k=1}^K{\lambda}_k{\theta_k}$, where $\lambda$ is the importance score.

Guaranteeing shard fairness alone is insufficient to achieve an unbiased final model due to the heterogeneity of data distributions across shard subgraphs, which limits the effectiveness of shard debiasing for the entire graph. Additionally, aggregating the posteriors from shard models introduces more bias, as the message passing used to obtain these posteriors amplifies and propagates the bias of a node to the neighbors. To address this, we propose \textit{global alignment}, which aims to reduce the disparity in prediction distributions across different sensitive groups, resulting in a model that is generally non-discriminatory to any group in the global graph.
\begin{align}
\label{eq:global_dp}
\fairloss_{global} = \vert P( \hat{\glabel}'\!=\!1|\senattri'\!=\!0) - P(\hat{\glabel}'=1|\senattri' = 1)\vert,
\end{align}
where $N$ is the number of nodes on $\graph'$, and and $N_k$ is the number of shard data considered.
\begin{align}
            &\totalloss_{global} =  \utilityloss_{global} + \alpha \fairloss_{global} ,         \label{eq:global_loss}\\
            &\totalloss_{k} =   
            \utilityloss_{k} +  \alpha_{k} \fairloss_{k} + \beta_k \totalloss_{global}, \label{eq:local_loss}
\end{align}
where $\utilityloss_{k}$ and $\fairloss_{k}$ can be obtained by \cref{eq:utility_loss} and \cref{eq:local_dp}. $\totalloss$ is the loss function. The final objective function is defined as follows:
\begin{align}
&\min _{\theta,\lambda} \totalloss_{k}\left(\model_{{\tilde{\theta}}}(\graph'),\glabel\right),
\\
&\text{ s.t. }  {\tilde{\theta}}\!=\! \sum_{k=1}^K{\lambda}_k{\theta_k}, \ \text{ and }
\theta_k=
\min _{\theta_k}\totalloss_{k}\left(\model_{\theta_k}({\graph'_k}),\glabel_k\right),
\end{align}
where $\tilde{\theta}$ is the aggregated model weights and will be used for predicting $\hat{\glabel}'$ on the entire graph and calculating the global fair loss.
And {\fgu} finally output the $K$ fair post-unlearning model 
$\theta=\{\theta_1,\cdots,\theta_K\}$ and the importance scores of the shard models $\lambda=\{\lambda_1,\cdots,\lambda_K\}$.

\noindent\textbf{Alternative Optimization.}
Simultaneously optimizing $\model_{k}'$ and $\lambda$ is challenging due to their interdependence. To address this, we employ an alternative optimization strategy, iteratively updating $\model_{k}'$ and $\lambda$ in distinct time periods. At each epoch, we first update the shard model $\model_{k}'$. During the initial $t_1$ training epochs, we also update $\lambda$. This iterative approach helps manage the complexity of their interdependence and improves overall optimization. This procedure is summarized as follows:
\begin{equation}
\begin{array}{c}
{\theta}_{k}^{t+1} = \theta_{k}^{t}-\eta \nabla_{\theta_{k}^t} \totalloss_{k} \  \text{(every epoch)},\\
\lambda^{t+1} = \lambda^{t}-\eta \nabla_{\lambda^{t}} \totalloss_{global} \ \text{(every $t_1$ epochs)}.
\end{array}\label{eq:retrainup}
\end{equation}
{\fgu} iterates this process until a predefined stopping condition is met, such as reaching a convergence criterion or completing a fixed number of epochs. The detailed algorithm is presented in~\cref{alg:overall_workflow}.

\begin{table*}[t]
    \centering
    \setlength{\tabcolsep}{5pt}
    \renewcommand{\arraystretch}{1.0}
            \caption{Utility and fairness on six datasets before (original training) and after unlearning at various node/edge unlearning ratios. We report mean $\pm$ std over $10$ runs. \hl{Cyan cells} exceed~\textit{fair retraining}, and the best among graph unlearning methods (\gedit, \ges, \gif, \gd, \fgu) are in \textbf{bold}.
}
\scalebox{1.0}{
    \begin{tabular}{lc| c c c | c c c | c c c}
    \toprule
    & Dataset& \multicolumn{3}{c|}{\german} & \multicolumn{3}{c|}{\credit} & \multicolumn{3}{c}{\pokecn}  \\
    \cline{2-11}
    &   Edge Unl. ($r_e$) & \multicolumn{3}{c|}{$10\%$} & \multicolumn{3}{c|}{$10\%$} & \multicolumn{3}{c}{$10\%$}\\
    \cline{2-11}
& Node Unl. ($r_n$) & $5\%$                                  & $10\%$                    & $20\%$                      & $5\%$                    & $10\%$                    & $20\%$                                         & $5\%$                    & $10\%$                   & $20\%$            \\
    \toprule
\multirow{7}{*}{\rotatebox{90}{Accuracy ($\uparrow$)}} 
& Retrain       &\ms{68.4}{0.9}                            &\ms{67.2}{1.5}             &\ms{65.7}{2.7}             &\ms{70.4}{2.1}              &\ms{69.3}{1.7}           &\ms{66.5}{2.2}                                &\ms{68.6}{1.0}              &\ms{66.6}{2.1}       &\ms{63.6}{1.9}   \\ 
& Fair Retrain  &\ms{67.1}{1.2}                            &\ms{67.1}{1.5}             &\ms{64.5}{2.2}             &\ms{70.1}{2.3}              &\ms{68.2}{1.3}           &\ms{65.2}{3.6}                                &\ms{68.1}{2.2}              &\ms{64.2}{1.7}       &\ms{60.2}{1.3}   \\\cmidrule(r){2-11}
& {\gedit}      &\ms{65.3}{2.0}                            &\underline{\ms{66.7}{2.1}} &\ms{62.8}{3.0}             &\ms{66.6}{1.7}              &\textbf{\ms{67.6}{2.0}}  &\ms{63.1}{1.7}                                &\ms{65.9}{1.5}              &\ms{62.6}{1.4}       &\ms{57.3}{1.0}  \\
& {\ges}        &\ms{66.2}{2.4}                            &\ms{65.1}{1.2}             &\underline{\ms{63.1}{1.4}} &\ms{65.4}{2.4}              &\ms{66.4}{1.6}           &\ms{61.6}{2.5}                                &\ms{64.5}{1.7}              &\ms{61.8}{2.2}       &\underline{\ms{58.6}{2.2}}  \\
& {\gif}        &\ms{66.3}{1.7}                            &\ms{64.2}{1.8}             &\ms{62.9}{2.3}             &\underline{\ms{66.9}{1.3}}  &\ms{65.9}{1.8}           &\cellcolor{LightCyan} \textbf{\ms{66.4}{3.2}} &\ms{65.3}{1.4}              &\ms{62.5}{1.8}       &\ms{56.5}{2.7}   \\    
& {\gd}         &\underline{\ms{66.4}{2.3}}                &\ms{65.1}{1.1}             &\ms{61.6}{1.5}             &\ms{64.3}{3.3}          &\ms{66.3}{1.8}           &\ms{64.3}{3.4}                                &\underline{\ms{66.6}{2.5}}  &\underline{\ms{63.3}{1.4}}   &\ms{55.8}{2.1}   \\    
& {\textbf{\fgu}}   &\cellcolor{LightCyan}\textbf{\ms{67.6}{1.8}} &\textbf{\ms{66.8}{1.5}} &\textbf{\ms{63.3}{1.2}} &  \textbf{\ms{68.5}{1.9}} &\underline{\ms{67.5}{2.3}}    &\cellcolor{LightCyan} \underline{\ms{65.6}{1.0}}  &\textbf{\ms{67.2}{2.2}}     &\textbf{\ms{64.1}{1.9}} &  \textbf{\ms{59.1}{1.4}}   \\    
\midrule             
\multirow{7}{*}{\rotatebox{90}{F1 ($\uparrow$)}} 
& Retrain       &\ms{79.1}{2.1}              &\ms{78.3}{1.2}              &\ms{76.5}{1.1}                &\ms{80.4}{1.3}              &\ms{79.6}{1.2}          &\ms{76.2}{2.3}          &\ms{63.3}{0.5}          &\ms{60.3}{0.8}                                  &\ms{58.7}{2.5}               \\ 
& Fair Retrain  &\ms{78.7}{1.7}              &\ms{77.4}{1.2}              &\ms{74.1}{2.4}                &\ms{79.2}{2.1}              &\ms{79.3}{0.9}          &\ms{77.1}{3.6}          &\ms{60.2}{2.5}          &\ms{59.1}{1.5}                                  &\ms{57.2}{2.3}              \\\cmidrule(r){2-11}
& {\gedit}      &\underline{\ms{77.3}{2.4}}  &\ms{75.4}{3.4}              &\underline{\ms{73.7}{1.9}}    &\underline{\ms{77.3}{1.4}}  &\textbf{\ms{78.2}{1.6}} &\ms{73.3}{2.1}          &\ms{58.6}{1.9}          &\cellcolor{LightCyan} \textbf{\ms{60.1}{2.4}}   &\ms{55.2}{1.6}              \\
& {\ges}        &\ms{75.0}{1.8}              &\ms{73.1}{2.5}              &\ms{71.9}{2.3}                &\ms{76.7}{1.5}              &\underline{\ms{78.1}{1.7}} &\ms{74.3}{3.3}       &\textbf{\ms{59.4}{1.6}} &\ms{58.2}{1.6}                                  &\underline{\ms{56.8}{2.7}}  \\
& {\gif}        &\ms{76.4}{2.2}              &\ms{75.1}{1.4}              &\ms{73.0}{2.7}                &\ms{76.9}{2.7}              &\ms{76.2}{1.5}          &\ms{72.5}{1.5}          &\ms{57.9}{2.1}          &\ms{57.3}{2.2}                                  &\ms{55.0}{1.9}               \\    
& {\gd}         &\ms{76.2}{1.6}              &\underline{\ms{76.6}{2.3}}  &\ms{72.8}{1.3}                &\ms{77.0}{2.1}              &\ms{76.0}{1.6}          &\underline{\ms{74.4}{2.4}} &\ms{58.3}{1.4}          &  \ms{58.9}{1.3}                                  &  \ms{56.1}{1.6}          \\    
& {\textbf{\fgu}}   &  \textbf{\ms{77.6}{1.5}}  &  \textbf{\ms{77.1}{2.1}} &  \cellcolor{LightCyan}\textbf{\ms{74.4}{1.6}} &  \textbf{\ms{78.5}{1.6}}  &\ms{77.6}{1.9}          &  \textbf{\ms{77.0}{1.4}} &\underline{\ms{59.1}{1.8}}          &  \cellcolor{LightCyan} \underline{\ms{60.0}{2.2}}            &  \cellcolor{LightCyan} \textbf{\ms{57.4}{2.1}}   \\    
\midrule
\multirow{7}{*}{\rotatebox{90}{$\Delta_{\mathit{DP}}$ ($\downarrow$)}} 
& Retrain       &\ms{19.8}{1.5}                       &\ms{32.7}{17.4}                &\ms{40.5}{26.7}                 &\ms{16.8}{6.2}                   &\ms{18.3}{11.4}                 &\ms{21.3}{22.5}               &\ms{12.3}{6.7}         &\ms{18.2}{5.3}                    &\ms{31.5}{16.7}  \\ 
& Fair Retrain  &\ms{2.4}{0.7}                        &\ms{4.3}{1.5}                  &\ms{6.8}{3.2}                   &\ms{5.4}{1.1}                    &\ms{6.7}{5.3}                   &\ms{4.6}{1.9}                 &\ms{2.5}{0.7}          &\ms{3.6}{1.7}                     &\ms{6.2}{2.3}   \\\cmidrule(r){2-11}
& {\gedit}      &\ms{27.1}{19.2}                      &\ms{45.1}{25.2}                &\ms{47.2}{39.6}                 &\ms{15.9}{7.7}                   &\ms{26.6}{26.0}                 &\ms{27.3}{15.6}               &\ms{21.1}{11.5}        &\ms{16.2}{8.3}                    &\ms{26.0}{19.4}  \\
& {\ges}        &\ms{31.2}{21.4}                      &\ms{47.1}{21.6}                &\ms{48.0}{21.0}                 &\ms{19.1}{7.5}                   &\ms{24.4}{17.6}                 &\ms{33.5}{12.7}               &\ms{22.1}{18.2}        &\ms{18.6}{7.3}                    &\ms{36.8}{21.0}  \\
& {\gif}        &\ms{23.7}{18.0}                      &\underline{\ms{44.3}{7.3}}     &\underline{\ms{32.8}{26.9}}     &\underline{\ms{13.5}{10.7} }     &\ms{15.8}{20.3}                 &\underline{\ms{22.3}{8.7}}    &\underline{\ms{16.5}{7.3}}    &\ms{14.3}{6.6}  &\underline{\ms{23.3}{22.1}}  \\    
& {\gd}         &\underline{\ms{18.9}{11.5}}          &\ms{46.1}{20.5}                &\ms{40.1}{22.1}                 &\ms{15.6}{8.3}                   &\underline{\ms{13.1}{19.5}}     &\ms{24.6}{23.1}               &\ms{19.3}{7.4}         &\underline{\ms{12.6}{6.4}}                    &\ms{29.4}{20.6}  \\    
& {\textbf{\fgu}}        &  \cellcolor{LightCyan}  \textbf{\ms{1.7}{0.6}}  &  \cellcolor{LightCyan} \textbf{\ms{2.1}{0.5}}   &  \cellcolor{LightCyan}  \textbf{\ms{3.9}{1.4}}      &  \cellcolor{LightCyan} \textbf{\ms{1.7}{1.2}}   & \cellcolor{LightCyan}  \textbf{\ms{3.7}{2.8}} &  \cellcolor{LightCyan} \textbf{\ms{1.5}{0.7}}   &  \cellcolor{LightCyan} \textbf{\ms{2.6}{2.1}}  &  \cellcolor{LightCyan} \textbf{\ms{2.1}{0.4}}  &  \cellcolor{LightCyan}  \textbf{\ms{3.2}{1.5}}    \\          
\midrule
\multirow{7}{*}{\rotatebox{90}{$\Delta_{\mathit{EO}}$ ($\downarrow$)}} 
& Retrain       &\ms{14.5}{6.2}                               &\ms{26.8}{10.2}                  &\ms{33.6}{20.1}                &\ms{8.6}{7.2}                                   &\ms{15.4}{8.7}                  &\ms{17.9}{10.3}                                  &\ms{19.9}{9.2}               &\ms{26.4}{18.5}              &\ms{31.2}{21.5} \\ 
& Fair Retrain  &\ms{3.5}{2.1}                                &\ms{2.4}{0.7}                    &\ms{3.3}{2.3}                  &\ms{14.5}{10.8}                                 &\ms{6.9}{3.4}                   &\ms{20.5}{17.1}                                  &\ms{3.0}{3.4}                &\ms{2.2}{4.3}                &\ms{6.5}{5.6}    \\\cmidrule(r){2-11}
& {\gedit}      &\ms{18.9}{12.6}                              &\ms{36.8}{9.3}                   &\ms{35.3}{17.5}                &\ms{16.3}{11.7}                                 &\ms{19.6}{8.7}                  &\ms{25.2}{12.4}            &\ms{13.2}{8.4}               &\ms{17.4}{7.1}              &\ms{21.8}{17.3}  \\
& {\ges}        &\ms{23.3}{21.7}                              &\ms{37.1}{19.7}                  &\ms{38.1}{24.2}                &\ms{19.8}{15.4}                                 &\ms{17.3}{5.6}                  &\ms{22.8}{25.0}                                  &\ms{15.5}{13.6}              &\ms{18.4}{8.9}               &\ms{29.7}{20.9}    \\
& {\gif}        &\ms{10.4}{5.3}                               &\underline{\ms{32.7}{6.8}}       &\ms{30.0}{16.7}                &\underline{\ms{10.6}{6.2}}                      &\ms{16.4}{5.8}                  &\underline{\ms{19.0}{6.3}} &\underline{\ms{7.6}{8.1}}    &\ms{14.6}{10.5}               &\ms{24.3}{12.5}  \\    
& {\gd}         &\underline{\ms{9.3}{4.7}}                    &\ms{33.5}{9.0}                   &\underline{\ms{29.2}{14.6}}    &\ms{11.2}{4.5}                                  &\underline{\ms{15.9}{7.6}}       &\ms{20.2}{8.5}             &\ms{10.5}{5.3}               &\underline{\ms{12.9}{6.4}}   &\underline{\ms{18.7}{7.2}}\\    
& {\textbf{\fgu}}        &  \cellcolor{LightCyan} \textbf{\ms{1.2}{0.8}} &  \cellcolor{LightCyan}\textbf{\ms{1.3}{0.4}}  &  \cellcolor{LightCyan}\textbf{\ms{2.1}{1.4}}  &\cellcolor{LightCyan}\textbf{\ms{1.1}{1.7}}  &\cellcolor{LightCyan}\textbf{\ms{2.3}{1.0}} & \cellcolor{LightCyan} \textbf{\ms{2.2}{1.8}}     &  \cellcolor{LightCyan} \textbf{\ms{2.1}{1.7}}  &  \cellcolor{LightCyan}\textbf{\ms{1.4}{0.6}}   &  \cellcolor{LightCyan} \textbf{\ms{1.6}{2.6}}    \\    
\bottomrule
\end{tabular} }  
        \label{tab:main_exp}
        \vspace{-1em}
    \end{table*}

\section{Experiment}
In our experiment, we focus on the node classification task within the graph unlearning setting. We validate the effectiveness of {\fgu} through three main types of experiments: utility, fairness, and privacy, under both node and edge unlearning. For utility performance, we measure the accuracy and F1 score of {\fgu} across various unlearning types and ratios. In terms of fairness, we examine the fairness-accuracy trade-off by evaluating whether {\fgu} achieves acceptable accuracy while maintaining a low $\Delta_{DP}$ compared to other graph unlearning baselines. Regarding privacy, we perform membership inference attacks to determine if the requested data has been effectively unlearned. 
\subsection{Experimental Setup}
\noindent\textbf{Datasets and Baselines.}
We conduct experiments using six fairness graph datasets from various domains, each with different sensitive attributes: {\german}, {\bail}, {\credit}, {\pokecn}, {\pokecz}, and {\nba}. To assess the efficacy and utility of our proposed method \fgu, we apply it to four state-of-the-art graph unlearning models: {\gedit}, {\ges}, {\gif}, and {\gd}.
Detailed information about the datasets can be found in~\cref{appendix: dataset_details}.

\noindent\textbf{Data Deletion Strategies.}\label{subsection:deletion_strategy}
To simulate real-world ``\textit{right to be forgotten}'' (RTBF) requests, we consider five types of data deletion requests for each dataset. These types combine different sensitive groups with different classes, encompassing four specific sensitive groups and one mixed group. For instance, if the sensitive attribute is \textit{gender}, the five groups are female positive, female negative, male positive, male negative, and a mixed group. To better reflect real-world deletion requests, we categorize them into three types: Privileged Group, Unprivileged Group, and Uniform Deletion. Detailed introductions about these requests are in~\cref{app:delete_req}.

\begin{figure*}[]
\centering
\includegraphics[width=1.0\linewidth]{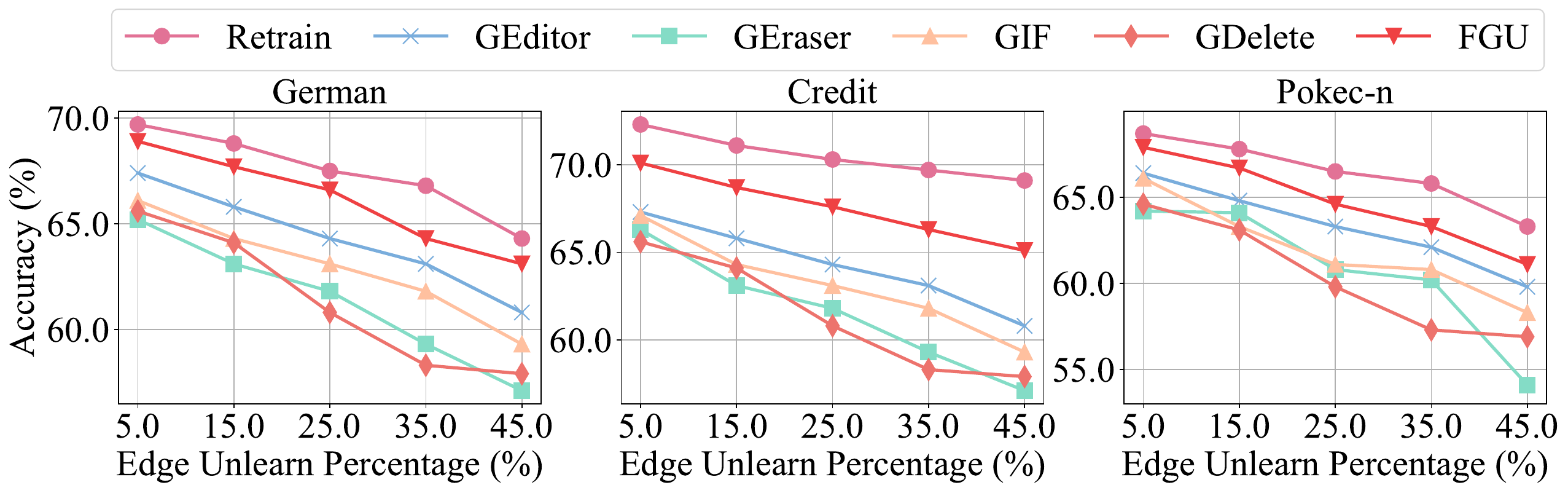}
\vspace{-1em}
\caption{Accuracy of graph unlearning methods on three datasets, under $10\%$ of node unlearning ratio and different edge unlearning ratios.} \label{exp:unlearning_rate}
\vspace{-1em}
\end{figure*}
\noindent\textbf{Metrics.} We employ two key evaluation metrics: prediction performance and fairness. For prediction performance, we measure accuracy and F1 score (the harmonic average of precision and recall), which is consistent with previous studies. To evaluate fairness, we utilize the Demographic Parity Difference $\triangle_{EO}$~\citep{zafar2017fairness, feldman2015certifying}, and the Equal Opportunity Difference $\triangle_{DP}$~\citep{hardt2016equality}, which are defined in~\cref{appendix:group_fair}. Detailed experimental setups are in~\cref{appendix:detail_setup}.
\subsection{Can FGU make fairer decisions compared to standard graph unlearning approaches?}
To evaluate whether graph unlearning introduces bias, we conducted tests on the fairness and utility performance of {\fgu}. ~\cref{tab:main_exp} provide the following observations:\\
\noindent\ballnumber{1}~\textbf{Both exact and approximate graph unlearning methods can introduce bias.} Generally, it is notable that the increase in $\triangle_{EO}$ and $\triangle_{DP}$ of the graph unlearning baselines compared to training on the original graph. For instance, on {\credit}, the $\triangle_{DP}$ is $2.25\times$ and $1.17\times$ higher in the exact graph unlearning approach {\gedit} and {\ges}, and $1.35\times$ and $1.83\times$ higher for approximate graph unlearning methods, namely {\gif} and {\gd}, respectively, compared to general training, as reported in~\cref{appendix:exp_original}. This suggests that unlearning nodes and edges in a graph introduces additional bias. This observation aligns with findings in \textit{machine unlearning} on tabular data as discussed in~\citep{Oesterling2023FairMU}, and our earlier preliminary analysis in~\cref{fig:intro}, indicating a universal phenomenon for unlearning data from a trained model.\\
\ballnumber{2}~\textbf{When increasing the node deletion ratio from $10\%$ to $20\%$, and then to $30\%$ on each dataset, the graph unlearning method exhibits an increased $\triangle_{EO}$ and $\triangle_{DP}$}, suggesting that more bias is introduced as more data is unlearned. This indicates that the bias is caused by the shift in data distribution among sensitive groups. In other words, increasing the deletion ratio leads to larger shifts and thus greater disparities in model predictions across different sensitive groups.
Particularly, the message passing of GNNs propagates the bias of a node to the neighbors, and the aggregated node representation will adopt more neighbors’ information with the same sensitive attribute, thus \textit{exaggerates the discrimination in the representation}~\citep{jiang2022fmp,qian2024addressing}. \\
\ballnumber{3}~\textbf{Exact graph unlearning methods generally exhibit more bias than approximate graph unlearning methods.} Specifically, {\gedit} and {\ges} have higher $\triangle_{EO}$ and $\triangle_{DP}$ than {\gif} and {\gd} on all datasets. This disparity can be attributed to the inherently \textit{aggressive} nature of exact graph unlearning methods. These methods \textit{directly delete} the requested data, which can inadvertently introduce more bias into the model.\\
\ballnumber{4}~\textbf{{\fgu} can achieve better fairness compared to graph unlearning baselines.}
The values of $\triangle_{EO}$ and $\triangle_{DP}$ for {\fgu} are lower than those of both retraining and graph unlearning baselines across all datasets. This observation suggests that {\fgu} is capable of achieving superior fairness, and in some cases, even matches the performance of fair retraining. This underscores the effectiveness of the fairness regularizer in {\fgu}, which guides shard retraining towards the goal of generating fair predictions. For instance, on the {\german} dataset, the $\triangle_{DP}$ of {\fgu} is only $5\%$, $6.0\%$, $5.8\%$, and $4.7\%$ of {\gedit}, {\ges}, {\gd}, and {\gif}, demonstrating its substantial improvement over graph unlearning baselines.\\
\ballnumber{5}~\textbf{The accuracy and F1 scores of {\fgu} surpass those of graph unlearning baselines and are comparable to fair retraining.} The experimental results show that, {\fgu} outperforms fair retraining baselines on datasets such as {\bail}, {\credit}, {\pokecz}, and {\nba}. The $\triangle_{EO}$ and $\triangle_{DP}$ metrics of {\fgu} closely align with those of fair retraining across all datasets. This suggests that {\fgu} achieves enhanced fairness without significant loss in utility. Consequently, developing {\fgu} is crucial, which demonstrates comparable fairness to fair retraining.

\subsection{Can FGU effectively forget the requested data?}
\begin{table}[t]
\caption{Attack accuracy of different edge unlearning ratios for graph unlearning, with edge unlearning ratio $10\%$. For the graph unlearning methods, the best results are in \textbf{bold}, and the runner-up results are underlined.
}
\label{exp:unlearning_mia}
\centering
\scriptsize
\setlength{\tabcolsep}{5pt}
\renewcommand{\arraystretch}{0.6}
\scalebox{1.0}{
\begin{tabular}{llc|ccccc} \toprule
Dataset& {$r_n$}            & {Retrain}   & {\textsf{GEditor}}& {\textsf{GEraser}} & {\textsf{GIF}} & {\textsf{GDelete}}& {\textbf{\textsf{FGU}}} \\\midrule
\multirow{3}{*}{{\german}} 
& \textbf{$5\%$}            & 51.2        & 63.1 &\underline{62.3}  & 66.7 & 64.5 & \textbf{51.2}  \\
& \textbf{$10\%$}           & 51.4        & 65.8 & \underline{64.5}  & 70.5 & 80.1 & \textbf{51.4}  \\
& \textbf{$20\%$}           & 51.9        & \underline{71.2} & 72.1  & 77.4 & 84.3 & \textbf{51.9}  \\
\midrule
\multirow{3}{*}{{\bail}} 
& \textbf{$5\%$}            & 51.0        & 64.6 & \underline{63.9}  & 68.9 & 70.8 & \textbf{51.5}  \\
& \textbf{$10\%$}           & 50.9        & 67.4 & \underline{66.3} &69.2 & 72.2   & \textbf{50.9}  \\
& \textbf{$20\%$}           & 51.2        & 78.0 & \underline{76.1} &81.4   & 79.2 & \textbf{51.3}  \\
\midrule
\multirow{3}{*}{{\credit}}   
& \textbf{$5\%$}           & 51.6        & \underline{71.5} & 72.5  &78.8  & 79.0 & \textbf{51.6}  \\
& \textbf{$10\%$}          & 51.2        & \underline{72.7} & 74.2  & 77.3 & 77.4 & \textbf{51.2}  \\
& \textbf{$20\%$}          & 51.2        & \underline{65.2} &65.3 & 69.1   & 68.9 & \textbf{51.5}  \\
\midrule
\multirow{3}{*}{{\pokecz}}   
& \textbf{$5\%$}            & 50.3        & 69.6 & \underline{67.1} & 71.7  & 75.0 & \textbf{50.6}  \\
& \textbf{$10\%$}           & 51.1        & \underline{67.5} & 68.3 & 77.4   & 82.9 & \textbf{51.1}  \\
& \textbf{$20\%$}           & 50.8        & \underline{71.6} &75.4 & 78.6   & 81.5 & \textbf{50.8}  \\
\midrule
\multirow{3}{*}{{\pokecn}}   
& \textbf{$5\%$}            & 51.8        & \underline{65.4} & 68.9 & 73.7  & 78.8 & \textbf{51.8}  \\
& \textbf{$10\%$}           & 51.2        & \underline{72.1} & 75.3 & 76.3   & 84.3 & \textbf{51.2}  \\
& \textbf{$20\%$}           & 50.7        & 80.4 & \underline{79.6} & 81.5   & 88.6 & \textbf{50.7}  \\
\midrule
\multirow{3}{*}{{\nba}}  
& \textbf{$5\%$}            & 51.5        & 68.5 & \underline{64.7} & 76.6   & 75.2 & \textbf{51.5}  \\
& \textbf{$10\%$}           & 51.7        & \underline{68.5} & 70.1 & 79.0   & 72.4 & \textbf{51.3}  \\
& \textbf{$20\%$}           & 50.4        & \underline{77.2} & 78.2 & 81.1   & 83.1 & \textbf{50.5}  \\
\bottomrule
\end{tabular}
}
\vspace{-1em}
\end{table}
To verify the effectiveness of the proposed \fgu, we conduct experiments on membership inference.

\noindent\textbf{Membership Inference:} Membership Inference (MI) is a privacy attack that seeks to infer whether a particular data point was included in the training dataset used to train a machine learning model. This becomes especially significant in the context of graph unlearning~\citep{chien2022efficient,chen2022graph,cheng2023gnndelete}, which aims to safeguard data privacy. The success of MI can thus serve as a relevant metric for evaluating the effectiveness of unlearning techniques~\citep{yeom2018privacy}.

\begin{figure*}[t]
\hspace{-1em}
\includegraphics[width=1.0\linewidth]{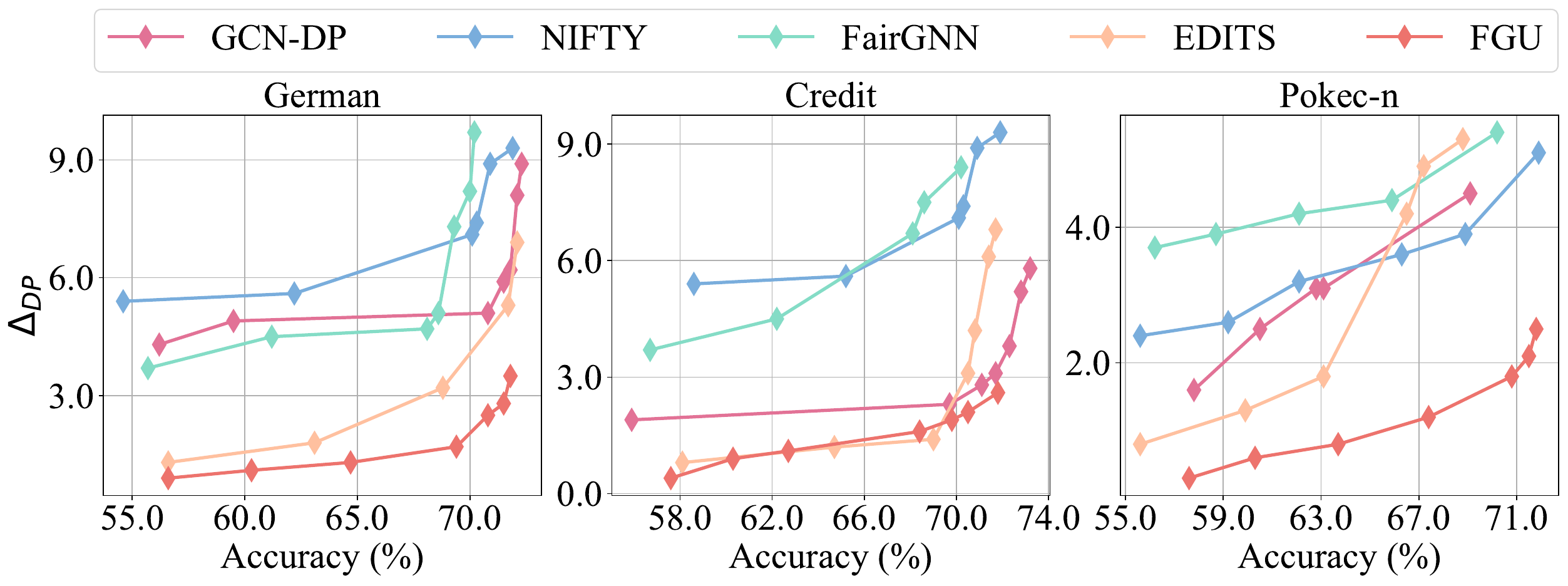}
\caption{Accuracy and $\triangle_{DP}$ trade-off comparison on three datasets and five fairness-aware GNNs. Results located in the bottom-right corner are preferable.}         \label{fig:tradeoff}
\vspace{-1em}
\end{figure*}    

\noindent\textbf{Setup:} 
In this study, we conduct MI attacks on the proposed graph unlearned algorithm, {\fgu}, as well as the standard retraining and graph unlearning baselines. These MIAs employ $20$ shadow models, and the target model has been trained for $100$ epochs. We evaluate the effectiveness of these MIAs based on the attack accuracy, the lower the attack accuracy, the more successful the unlearning, i.e., the MIA cannot infer whether the data is in the training set or not. We conduct $10$ runs on different and random splits and report the average results. 
From~\cref{exp:unlearning_mia}, we have the following observations:\\
\noindent\ballnumber{6}~\textbf{Exact graph unlearning methods generally achieve better privacy than approximate graph unlearning approaches.} This is because exact graph unlearning methods, such as {\gedit}, {\ges}, and the proposed {\fgu}, directly eliminate the impact of deleted data from the learned model. In contrast, approximate graph unlearning methods like {\gif} and {\gd} attempt to make the model weights indistinguishable from those trained on the remaining data. However, the limitation of approximate graph unlearning methods is that attackers may still recover some information from the post-unlearning model, as in previous studies~\citep{xu2023machine}.\\
~\ballnumber{7}~\textbf{{\fgu} can achieve comparable privacy performance with retraining.} The attack accuracy of {\fgu} is close to $50\%$, which is a random guess, indicating that the MIA cannot infer whether the data is in the training set or not. This shows that {\fgu} can achieve comparable privacy performance with retraining, and better than the graph unlearning baselines.\\
~\ballnumber{8}~\textbf{The increased node deletion ratio has no effect on the privacy of {\fgu} but affects the graph unlearning baselines.} The attack accuracy of {\fgu} is stable across different node deletion ratios, while the graph unlearning baselines have a significant increase in the attack accuracy with the increased node deletion ratio, which indicates that more unlearning will leak more data membership information.

\subsection{Can FGU perform stably across different deletion ratios and data distributions?}
To systematically evaluate the impact of different deletion request types on model performance, we conduct experiments where we simulate deletion requests by randomly sampling unlearned data from corresponding groups or the entire training dataset for the mixed group at specified ratios. We vary deletion ratios from $5\%$ to $45\%$ to assess model behavior under different levels of data removal.

From the results in~\cref{exp:unlearning_rate}, we observe:
\ballnumber{9} \textbf{{\fgu} demonstrates comparable accuracy with retraining and superior accuracy compared to graph unlearning baselines across different unlearning ratios}, highlighting its stability in task prediction. {\fgu} consistently maintains satisfactory accuracy and outperforms graph unlearning baselines, although the accuracy gap between {\fgu} and retraining increases with higher unlearning ratios. The widening accuracy gap may result from the increased introduction of bias as the unlearning ratio rises, as depicted in~\cref{tab:main_exp}. Consequently, optimizing data with heightened bias becomes more challenging, given the necessity for fairness in {\fgu}, which in turn entails a greater sacrifice on utility.
In addition, we also conduct experiments on different unlearning types in~\cref{appendix:delete_distr}.

\subsection{How does the accuracy-fairness trade-off of FGU compare to fairness-aware GNN baselines? }
To evaluate the accuracy-fairness trade-offs of various GNN debiasing methods, namely GCN-DP~\citep{li2019kernel}, NIFTY~\citep{agarwal2021towards}, FairGNN~\citep{dai2021say}, and EDITS~\citep{dong2022edits}, we conducted experiments. Each debiasing method was applied to the remaining data obtained by directly removing the requested data from the input graph. The Pareto front curves~\citep{yao2023stochastic,ling2022learning} resulting from a grid search of hyperparameters for each method are depicted in \cref{fig:tradeoff}. It presents the accuracy versus $\Delta_{DP}$ for the baselines and {\fgu} across three datasets. The ideal performance, represented by the bottom-right corner point, corresponds to the highest accuracy and the lowest bias. We discover that:~\ballnumber{\tiny{10}}~\textit{{\fgu} $\Delta_{DP}$ and accuracy demonstrates a balanced trade-off between fairness and utility}, surpassing direct training of GNN debiasing methods on the remaining data. Specifically, {\fgu} achieves a superior accuracy-$\Delta_{DP}$ trade-off compared to all fairness-aware baselines across three datasets. When accuracy is held constant, {\fgu} exhibits lower $\Delta_{DP}$ compared to the baselines. Conversely, when mitigating the same level of $\Delta_{DP}$, {\fgu} achieves higher accuracy. These results affirm the efficacy of two-level debiasing of {\fgu}, which incorporates shard debiasing and global alignment in fair shard retraining.

\subsection{Can FGU balance the accuracy and fairness trade-off?}\label{appendix: hyperparameterr}
\begin{figure}[t]
    \centering
    \vspace{-0.5em}
    \hspace{-1.5em}
    \subfloat[\german]{\includegraphics[height=1.4in]{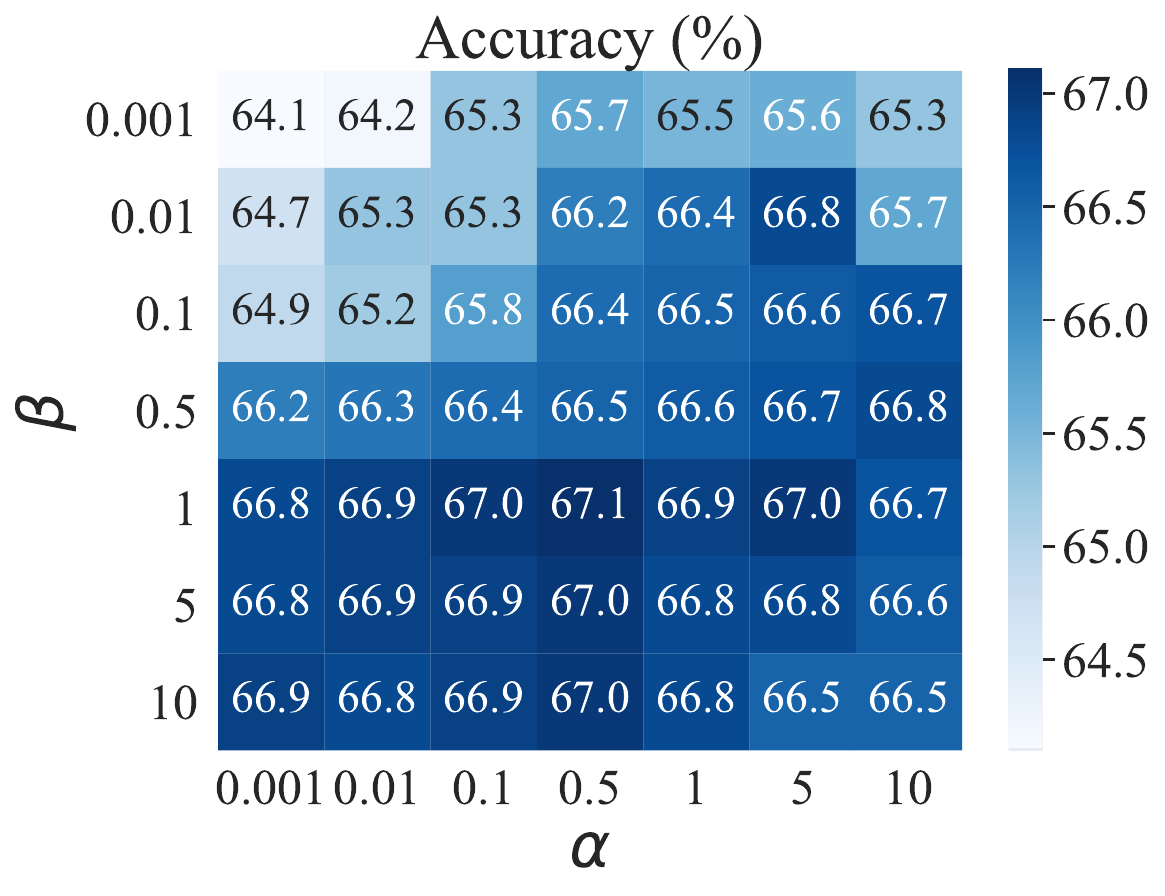}\includegraphics[height=1.4in]{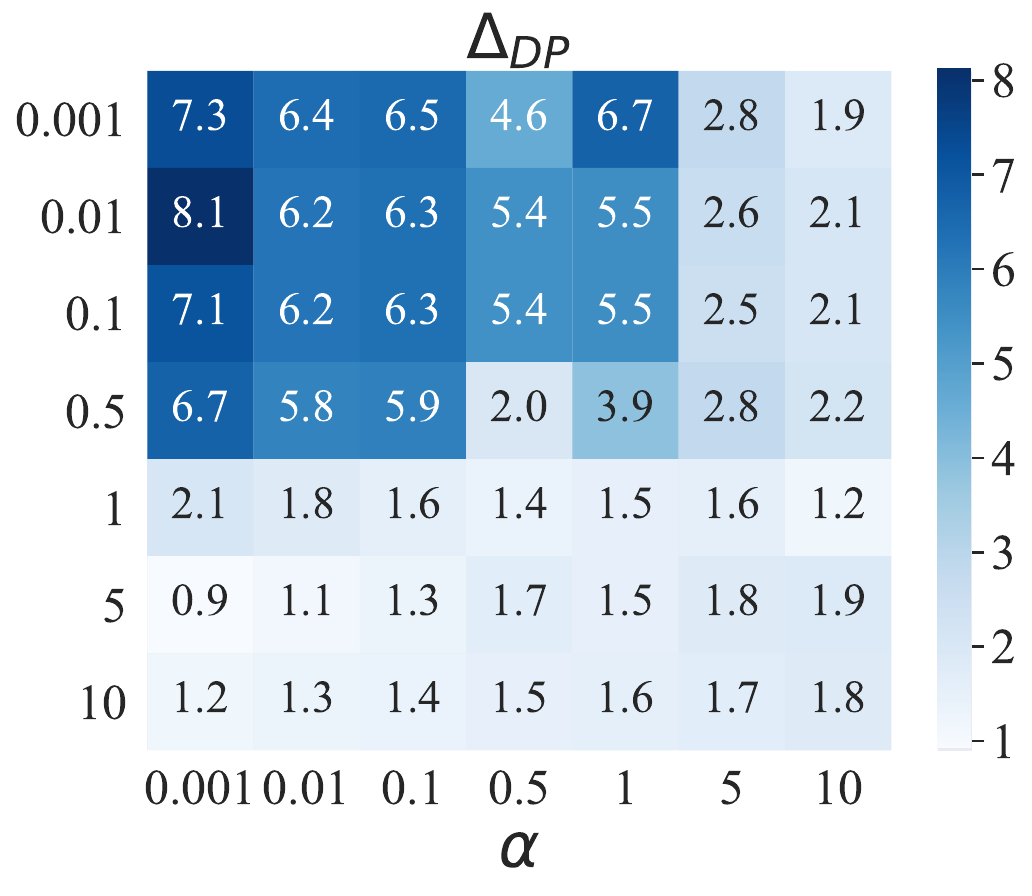}}\\
    \centering            
    \hspace{-1.5em}
    \subfloat[\credit]
    {\includegraphics[height=1.4in]{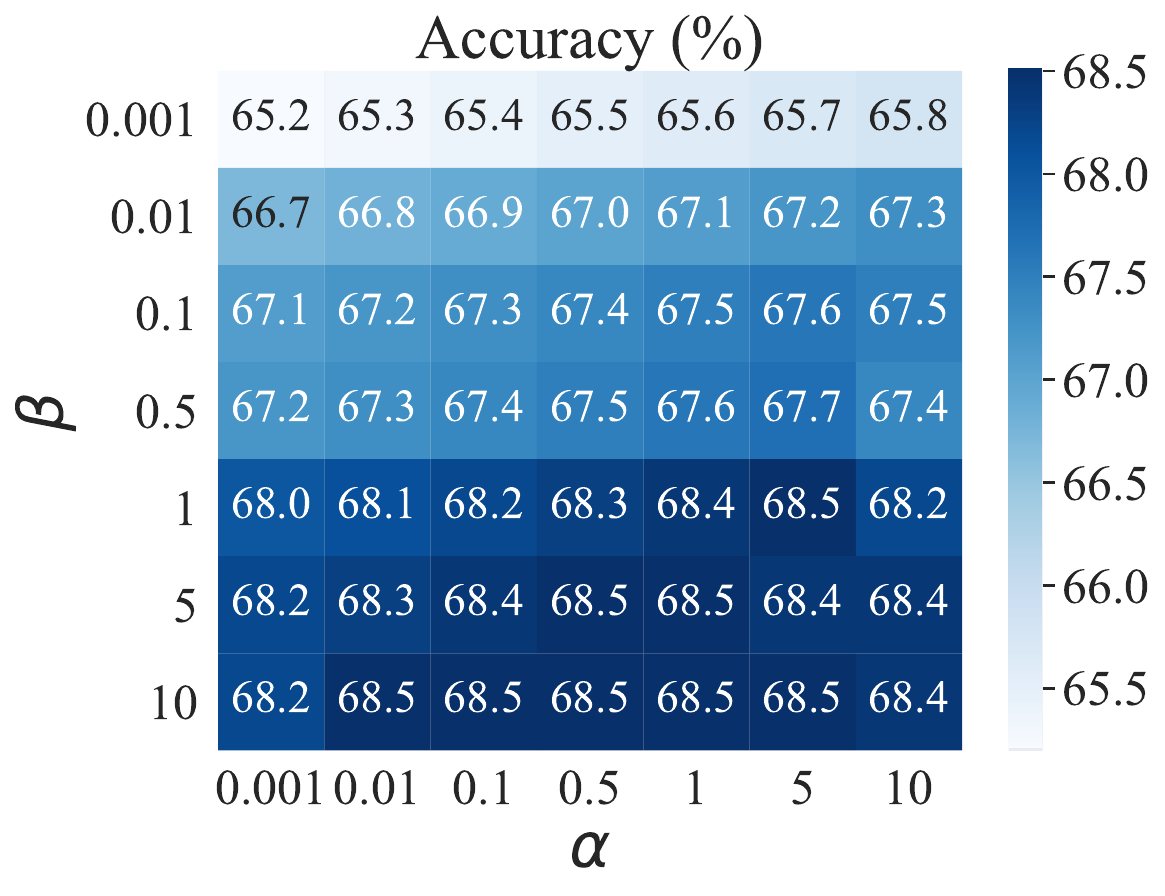}\includegraphics[height=1.4in]{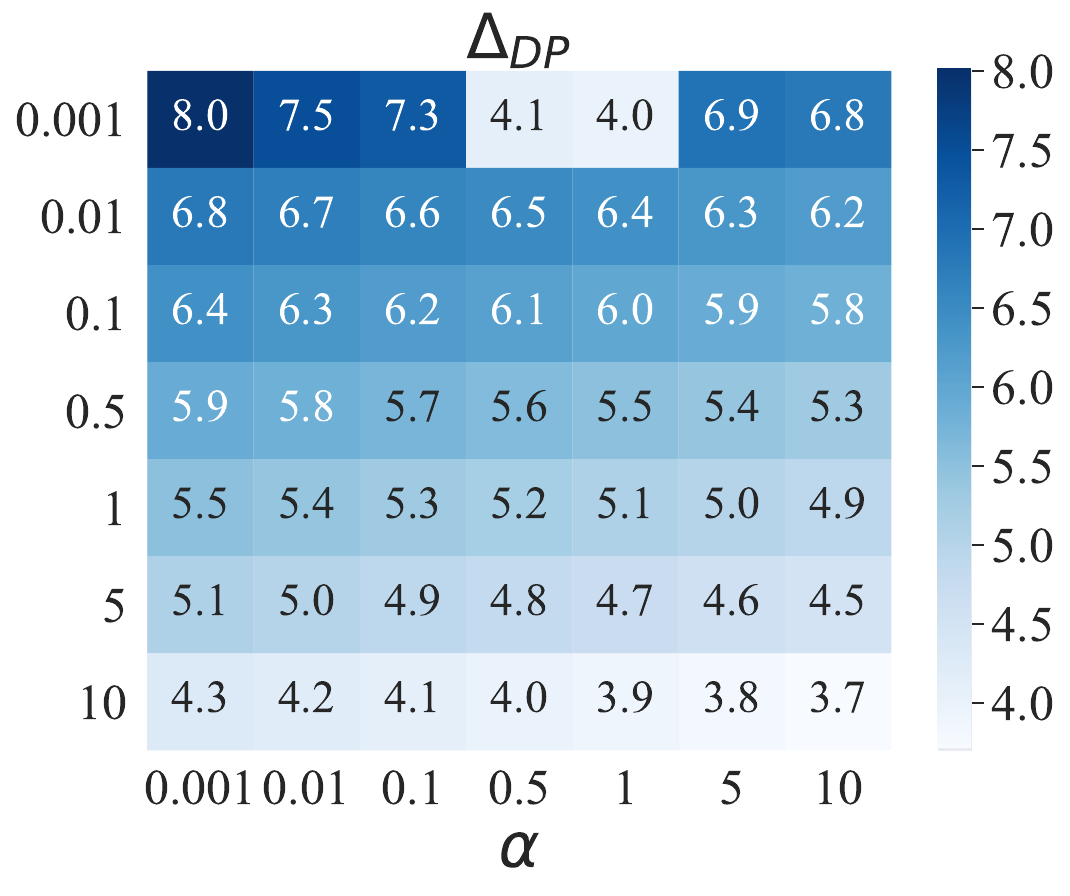}}\\
        \centering            
    \hspace{-1.5em}
            \subfloat[\pokecn]{\includegraphics[height=1.4in]{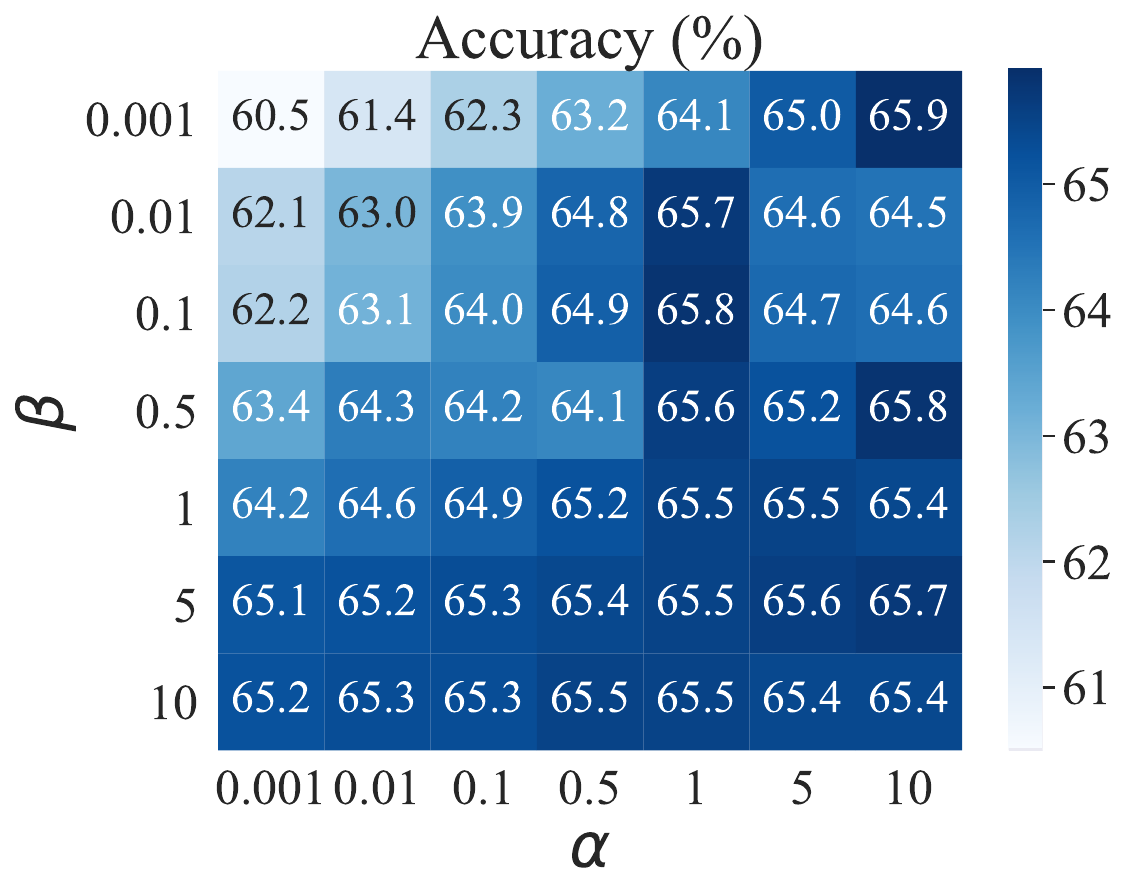} 
        \includegraphics[height=1.4in]{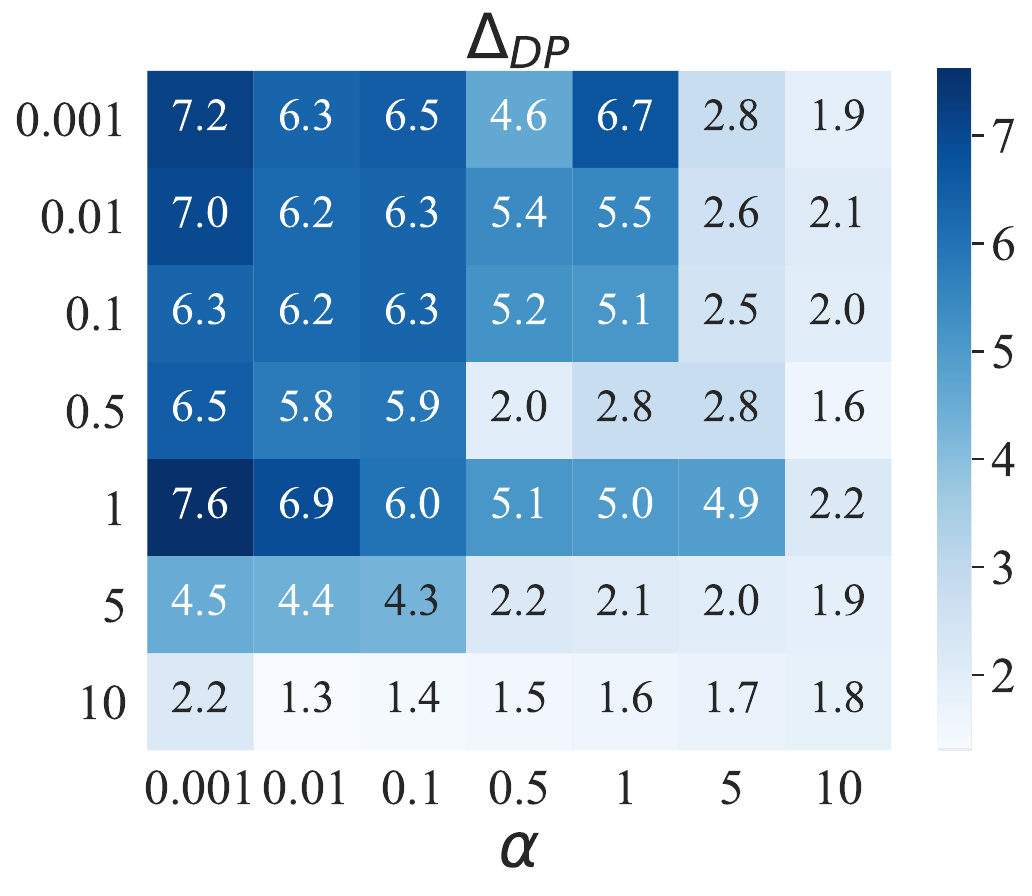}}
        \label{fig:hp_poken}
    \caption{Accuracy and $\triangle_{DP}$ trade-off on {\german}, {\credit}, and {\pokecn}. Results located in the bottom-right corner are preferable.}
    \label{fig:hp}
    \vspace{-1.5em}
\end{figure}
\begin{table*}[t]
\centering
\caption{The performance of standard training and fair training on the original graph. $\uparrow$ represents the larger the better, while $\downarrow$ represents the opposite.}\label{appendix:exp_original}
\resizebox{\textwidth}{!}{%
\begin{tabular}{ccccccccccccc}
\toprule
\multicolumn{1}{c|}{Dataset} & \multicolumn{2}{c|}{\german}            & \multicolumn{2}{c|}{\bail}             & \multicolumn{2}{c|}{\credit}           & \multicolumn{2}{c|}{\pokecz}          & \multicolumn{2}{c|}{\pokecn}          & \multicolumn{2}{c}{\nba} \\ \midrule
\multicolumn{1}{c|}{Metric}  & Standard   & \multicolumn{1}{c|}{Fair} & Standard  & \multicolumn{1}{c|}{Fair} & Standard  & \multicolumn{1}{c|}{Fair} & Standard  & \multicolumn{1}{c|}{Fair} & Standard  & \multicolumn{1}{c|}{Fair} & Standard   & Fair       \\ \midrule
\multicolumn{1}{c|}{F1($\uparrow$) }      & \ms{80.0}{1.5}  & \ms{82.1}{0.2}                 & \ms{79.4}{1.3} & \ms{79.5}{1.2}                 & \ms{82.9}{0.7} & \ms{82.6}{1.1}                 & \ms{67.2}{0.5} & \ms{67.1}{0.7}                 & \ms{65.2}{0.5} & \ms{63.2}{1.6}                 & \ms{74.4}{1.2}  & \ms{74.5}{0.7}  \\
\multicolumn{1}{c|}{ACC($\uparrow$) }     & \ms{71.5}{1.2}  & \ms{69.8}{0.6}                 & \ms{84.5}{1.2} & \ms{84.5}{1.1}                 & \ms{73.6}{0.8} & \ms{73.4}{1.2}                 & \ms{69.4}{0.2} & \ms{66.0}{2.5}                 & \ms{70.2}{0.4} & \ms{70.4}{2.2}                 & \ms{72.0}{0.7}  & \ms{70.3}{0.5}  \\
\multicolumn{1}{c|}{AUC($\uparrow$) }     & \ms{73.8}{2.1}  & \ms{69.5}{1.6}                 & \ms{88.8}{1.3} & \ms{89.1}{1.5}                 & \ms{67.7}{0.3} & \ms{69.0}{0.2}                 & \ms{74.9}{0.9} & \ms{70.4}{2.2}                 & \ms{74.9}{0.2} & \ms{73.8}{2.1}                 & \ms{77.0}{0.2}  & \ms{76.3}{0.5}  \\
    \multicolumn{1}{c|}{$\Delta_{\mathit{DP}}$($\downarrow$) }        & \ms{36.7}{11.6} & \ms{1.8}{3.2}                  & \ms{7.4}{1.1}  & \ms{3.2}{0.9}                  & \ms{11.5}{2.3} & \ms{4.4}{2.7}                  & \ms{4.8}{1.0}  & \ms{2.4}{1.5}                  & \ms{7.8}{0.8}  & \ms{5.8}{3.1}                  & \ms{2.0}{0.9}   & \ms{1.9}{1.3}   \\
\multicolumn{1}{c|}{$\Delta_{\mathit{EO}}$($\downarrow$)}        & \ms{28.8}{9.5}  & \ms{2.0}{3.1}                  & \ms{4.3}{1.0}  & \ms{2.8}{0.9}                  & \ms{9.6}{1.6}  & \ms{3.0}{2.0}                  & \ms{5.1}{1.1}  & \ms{2.2}{1.2}                  & \ms{11.6}{1.1} & \ms{7.6}{3.3}                  & \ms{3.3}{1.5}   & \ms{1.6}{2.1}   \\ \bottomrule
\end{tabular}
}
\end{table*}
This section investigates the influence of the hyperparameters in our proposed method {\fgu}. One important hyperparameter in {\fgu} is $\alpha$ and $\beta$, controlling the influence of the fairness regularizer on the model prediction.
To investigate the parameter sensitivity and find a good trade-off of achieving high accuracy with low $\triangle_{DP}$, we train {\fgu} on all datasets with various $\alpha$ values. More specifically, we alter the values of $\alpha$ and $\beta$ among $\{0.5, 1.5, 3.0, 5.0, 7.0\}$. The results are presented in \cref{fig:hp}.
Specifically, 
we observe that when $\alpha\le1.5$, the classification performance is almost unaffected. Once $\alpha$ is too large, the classifier’s performance will decay sharply. 
~\textbf{When increasing the value of $\alpha$, {\fgu} will first have decreased $\triangle_{DP}$, while when the value is too large, the fairness drops} because it would be difficult to optimize the model to the global minimum. Therefore, to achieve a balance of accuracy and fairness, in all other experiments, we choose $\alpha=3.0$ and $\beta=1.5$ to perform {\fgu}.

\subsection{What Are the Effects of Global and Local Debiasing Modules?}
In the design of {\fgu}, we conduct two ablation studies: one focusing on the global and local debiasing modules, and the other on different debiasing techniques, to assess their impact on the performance of {\fgu}.

We conducted an ablation study on the {\german} dataset. In this study, \textit{Global Debiasing} refers to the FGU variant that uses only global debiasing, optimizing for the global objective. \textit{Local Debiasing} refers to the method that optimizes only for the local objective. Other parts of the model, such as partitioning and aggregation, remain unchanged. 
\begin{table}[H]
\centering
\caption{Utility and fairness performance on the German dataset. Local Debiasing refers to optimizing fairness only at the local (shard) objective, whereas Global Debiasing applies fairness constraints solely during the aggregation process. }
\setlength{\tabcolsep}{4pt}
\renewcommand{\arraystretch}{1.0}
% \label{tab:ablation}
\begin{tabular}{lccc}
\toprule
Method     		                 & Global Debiasing  		 & Local Debiasing  		    & {\fgu}    \\ \midrule
ACC ($\uparrow$)                 & \ms{64.3}{1.7}            & \ms{77.1}{2.2}   		& \ms{66.8}{1.1} \\
F1 ($\uparrow$)	                 & \ms{73.5}{2.5}  			 & \ms{89.8}{0.4}    		& \ms{77.1}{2.1}  \\
$\triangle_{EO}$ ($\downarrow$)  & \ms{8.6}{4.3}  			 & \ms{3.2}{2.9} 			& \ms{2.9}{0.3}   \\
$\triangle_{EO}$ ($\downarrow$)  & \ms{5.4}{2.8}  			 & \ms{3.1}{1.6}            & \ms{2.8}{0.2} \\
\bottomrule
\end{tabular}
\label{exp:glo_lo}
\end{table}
We observe from~\cref{exp:glo_lo} that the proposed FGU performs better than its two variants. 

\subsection{How Does FGU Perform with Different Debiasing Techniques?}
We conducted additional experiments by combining other debiasing methods with {\fgu}. Specifically, we incorporated adversarial debiasing (Adv)~\citep{zhang2018mitigating}, PrejudiceRemover (PR)~\citep{kamishima2012fairness}, and LAFTR~\citep{madras2018learning}:
\begin{table}[]
\centering
\caption{The utility and fairness performance of intuitive methods across five datasets. Here, FGU${ADV}$, FGU${PR}$, and FGU$_{LAF}$ denote the ADV, PR, and LAFTR methods, respectively.}
\setlength{\tabcolsep}{4pt}
\renewcommand{\arraystretch}{1.1}
\label{tab:ablation2}
\begin{tabular}{lcc|cc}
\toprule
Method     		  & ACC ($\uparrow$)  		& F1 ($\uparrow$)  			& $\Delta_{DP}$ ($\downarrow$) 	& $\Delta_{EO}$ ($\downarrow$)    \\ \midrule
 {\fgu}$_{ADV}$  & \ms{\textbf{67.1}}{1.4}  			&  \ms{76.3}{1.3}   			& \ms{4.0}{3.3}   					& \ms{4.5}{3.7}     \\
 {\fgu}$_{PR}$ 	 & \ms{65.9}{0.7}  			& \ms{76.5}{1.2}    			& \ms{3.8}{2.6}    					& \ms{2.4}{1.2}     \\
 {\fgu}$_{LAF}$ 	 & \ms{66.6}{1.3}  			& \ms{76.8}{0.6} 			& \ms{2.5}{2.0}    					& \ms{1.9}{1.6}     \\
 {\fgu}              	 & \ms{66.8}{1.1}  			& \ms{\textbf{77.1}}{2.1}   			& \ms{\textbf{2.1}}{0.5}   					& \ms{\textbf{1.3}}{0.4}   \\
\bottomrule
\end{tabular}
\end{table}
From the results reported in~\cref{tab:ablation2}, we observe that FGU, which incorporates the DP regularizer, achieves a balanced trade-off between utility and fairness compared to the other three variants. Specifically, {\fgu}$_{ADV}$ shows strong utility but has the lowest fairness, while {\fgu}$_{LAF}$ performs well in both utility and fairness. These findings underscore the advantages of the DP regularizer and highlight the flexibility of the proposed bi-level debiasing method. This demonstrates its potential applicability to other in-processing debiasing methods~\citep{dai2022comprehensive}.

In conclusion, integrating the DP regularizer into graph unlearning effectively balances utility and fairness, making it a promising approach for achieving fairness.

\subsection{Complexity and Storage Analysis}
We conducted a runtime comparison between {\fgu} and the established baselines. The experiments were executed using NVIDIA RTX A4000 GPUs with 16GB GDDR6 Memory, with a fixed number of requests set at $100$. For both tabular and image datasets, we configured the retraining epoch to $100$. It's important to note that for {\gedit}, the number of shards~\citep{bourtoule2021machine} is set to $S=20$.
\begin{table}[H]
\centering
\caption{Runtime comparison (in seconds) on different graph unlearning methods.}
\setlength{\tabcolsep}{3pt}
\label{tab:runtime_comparison}
\begin{tabular}{lccccc}
\toprule
                    & Retrain & {\gedit}  & {\gif} & {Fair Retrain} & {\fgu}      \\ \midrule
Runtime ({\german})     & 103.5 s     & 25.5 s & 7.4 s    & 6.2 s      & 3.1 s    \\
Runtime ({\bail})  & 968.3 s   & 59.1 s & 5.6 s    & 22.2 s     & 4.2 s   \\
\bottomrule
\end{tabular}
\end{table}
Based on the results in~\cref{tab:runtime_comparison}, the following observations can be made:
\begin{itemize}[leftmargin=0.35cm, itemindent=.0cm, itemsep=0.0cm, topsep=0.0cm]
    \item The computational time for {\fgu} is consistently less than $2.1$ seconds and $5.2$ seconds on {\german} and seconds on {\bail}. This represents a significant improvement, being $38.2\times$ and $318.7\times$ faster than the retraining approach, respectively.
    \item Both Amenisac and {\fgu} demonstrate superior time efficiency when compared to retraining and {\gedit}. Notably, {\gedit} necessitates the retraining of sub-models, whereas Amenisac and {\fgu} operate without such a requirement.
    \item {\fgu} outperforms fair retraining in terms of time efficiency, owing to the limited retraining.
\end{itemize}
    
\section{Conclusion}
In this paper, we address the challenge of bias introduced by current graph unlearning approaches when removing requested data. To tackle this issue, we introduce {\fgu}, a novel framework employing two main strategies: shard debiasing and global alignment. Shard debiasing targets discrimination in predictions among sensitive groups, while global alignment ensures fairness in the aggregated distribution by adjusting model weights through back-propagation. Our extensive experiments across six datasets validate the effectiveness of {\fgu}, showcasing its superior performance in terms of both fairness and utility compared to existing graph unlearning methods. We systematically investigate the impact of different unlearning ratios and distributions on {\fgu}. In the future, we aim to develop even more efficient and effective fair graph unlearning methods tailored for graph classification, individual fairness, and feature unlearning.

\section*{Acknowledgments}
This work was supported in part by the DARPA Young Faculty Award, the National Science Foundation (NSF) under Grants \#2127780, \#2319198, \#2321840, \#2312517, and \#2235472, the Semiconductor Research Corporation (SRC), the Office of Naval Research through the Young Investigator Program Award, and MURI grant. Additionally, support was provided by the Air Force Office of Scientific Research under Award \#FA9550\-22\-1\-0253 and Army Research Office Grant \#W911NF2410360.

\bibliographystyle{IEEEtran}
\bibliography{ref}

\appendix

\section{Appendix}
\subsection{Additional Preliminaries}
\subsubsection{Details about Graph Neural Network}\label{appendix:gnn}
Here, we only introduce the design of GCN~\citep{kipf2016semi}, which is one of the most popular GNN architectures~\citep{liu2024tinygraph,zhou2020graph,liu2025graph}. 
More specifically, each layer of GCN can be written as:
\begin{equation}
    \mathbf{H}^{(k)} = \sigma(\tilde{\mathbf{A}}\mathbf{H}^{(k-1)}\mathbf{W}^{(k)}),\label{eq:GCN_layer}
\end{equation}
where $\mathbf{H}^{(k)}$ denote the representations of all the nodes after the $k$-th layer; $\mathbf{W}^{(k)}$ stands for the  parameters of the $k$-th layer. $\tilde{\mathbf{A}}$ is the normalized adjacency matrix. Generally, the symmetric normalized form is used, which can be written as $\tilde{\mathbf{A}}=\mathbf{D}^{-\frac{1}{2}}(\mathbf{A}+\mathbf{I})\mathbf{D}^{-\frac{1}{2}}$, and $\mathbf{D}$ is a diagonal matrix with $D_{ii}=\sum_{i}A_{ij}$. $\mathbf{I}$ is the identity matrix. $\sigma$ is the activation function such as ReLU, i.e, $\sigma(x) = \max(0, x)$.

\noindent\textbf{Definition of Node Classification Task.}
Given a training graph $\graph$ and a set of labels $\glabel$, the node classification task aims to learn a representation vector $h_v$ for $j$ using a learning model $\model$ such that $v'$s label can be predicted as $y_v = \model(h_v)$. 

\begin{equation}
    \hat{\glabel}  = 
    \begin{cases} 
      1 & \text{if } \mathbf{H}^{(L)} - 0.5 \geq 0 \\
      0 & \text{if } \mathbf{H}^{(L)} - 0.5 < 0 
    \end{cases}
    \end{equation}
where, $\sigma=\text{sigmoid}(x)=\frac{1}{1 + \exp(-x)}$.
\begin{equation}
        \hat{\glabel}=\sigma(\tilde{\mathbf{A}}\mathbf{H}^{(L-1)}\mathbf{W}^{(L)}),\\
        \operatorname{Pr}(\hat{\glabel}=1)=\frac{\sum{\hat{\glabel}}}{N}
\end{equation}
where $\tilde{\mathbf{A}}$ is obtained by $A$.
For simplicity, the GCN framework can be denoted as $\hat{\glabel}=GCN(A,X)$.

\subsection{Related Work of Graph Unlearning}\label{appendix:graph_unlearning}
The graph unlearning baselines can be categorized into exact graph unlearning and approximate graph unlearning. 
In our experiment, we compare the proposed {\fgu} with the exact graph unlearning methods, including the {\gedit} and {\ges}, aiming to remove the impact of deleted data from the learned model without fully retraining the model. In contrast, approximate graph unlearning methods, including the {\gif} and {\gd}, focus on approximating the unlearning process to reduce the computational burden associated with retraining. The following are the graph unlearning baselines considered in our experiments:
\begin{itemize}[leftmargin=0.35cm, itemindent=.0cm, itemsep=0.0cm, topsep=0.0cm]
\item\textbf{Retrain (Fair)} is conducted by integrating a fairness regularizer into the standard retraining process in our experiment. The standard retraining, referred to as 'Retraining (Standard)' in this paper, is widely recognized as the benchmark for graph unlearning literature~\citep{zhang2023review, xu2023machine, nguyen2022survey}. It involves retraining the model on the remaining data following the removal of the requested information from the original dataset. It preserves privacy performance and serves as the ceiling performance in comparison. However, retraining comes with a significant cost, motivating the exploration of alternative graph unlearning techniques.
\item\textbf{{\ges}} ({\grapheraser})~\citep{chen2022graph} stands as one of the pioneering works in the emerging domain of graph unlearning, providing a novel framework for addressing unlearning requests in graph machine learning models. This approach begins by partitioning the original training graph into separate and disjoint shards, and learns an optimal importance score for each shard model. The importance scores are then used to aggregate the shard models into a single model. When the unlearning request comes, {\ges} removes the data from the corresponding shards and then retrains the shard model. The {\grapheraser} approach is designed to protect the privacy of the data subjects by ensuring that the revoked data is forgotten by the model.
\item\textbf{{\gif}} (Graph Influence Function) aims to model the influence of each training data point on the model with respect to various performance criteria and subsequently eliminates the negative impact. The method incorporates an additional loss term that considers the influence of neighbors and estimates parameter changes in response to a $\epsilon$-mass perturbation in deleted data.
\item\textbf{{\gd}} formalizes two essential properties for the GNN deletion method: the deleted edge consistency and neighborhood influence. The deleted edge consistency ensures that the predicted probabilities for deleted edges in the post-unlearning model stay similar to those for nonexistent edges. Similarly, the neighborhood influence guarantees that predictions in the shard vicinity of the deletion retain their original performance and remain unaffected by the removal. In pursuit of efficiency and scalability, {\gd} employs a layer-wise deletion operator to modify a pre-trained GNN model. Upon receiving deletion requests, {\gd} freezes the existing model weights and introduces small, shared weight matrices across the nodes in the graph. {\gd} guarantees strong performance by ensuring that the difference between node representations obtained from the trained model and those obtained from {\gd} remains theoretically bounded.
\end{itemize}

\subsection{More Details on Experimental Settings}
\noindent\textbf{Dataset Details.}\label{appendix: dataset_details}
\begin{table*}[t]
\normalsize
\centering
\caption{Statistics of commonly-used semi-synthetic and real-world datasets in fair graph learning.}
\label{tab:dataset}
\setlength{\tabcolsep}{2pt}
\scalebox{0.85}{
\begin{tabular}{@{}lcccccc@{}}
\toprule
 & \textbf{\german} & \textbf{\bail} & \textbf{\credit} & \textbf{\pokecz} & \textbf{\pokecn} & \textbf{\nba} \\ \midrule
$\vert \nodeset \vert$ & 1,000 & 18,876 & 30,000 & 67,797 & 66,569 & 403 \\
$\vert \edgeset \vert$ & 21,742 & 311,870 & 1,421,858 & 617,958 & 517,047 & 10,621 \\
$\vert \feat \vert$ & 27 & 18 & 13 & 69 & 69 & 39 \\
Avg. Degree & 44.48 & 34.04 & 95.79 & 19.23 & 16.53 & 53.71 \\
Sens. & Gender & Race & Age & Region & Region & Nationality \\
Task & Credit Status & Bail Decision & Future Default & Working Field & Working Field & Salary \\ \bottomrule
\end{tabular}}
\end{table*}
Here we present a detailed description of the six widely used datasets we used to validate our proposed issues as follows:
\begin{itemize}[leftmargin=0.35cm, itemindent=.0cm, itemsep=0.0cm, topsep=0.0cm]
\item \textbf{\bail}: Nodes represent defendants released on bail between $1990$ and $2009$, with edges indicating similar criminal records and demographic characteristics. The goal is to predict bail status based on race as the sensitive attribute.

\item \textbf{\credit}: Nodes symbolize credit card users, connected by edges denoting similarity in purchasing and payment behaviors. The task is to identify users likely to default on payments, with age serving as the sensitive attribute.

\item \texttt{Pokec}: This dataset from the Slovak social network Pokec, anonymized in 2012, includes two subsets: \pokecz and \pokecz, representing user profiles from two significant regions. The goal is to predict users' employment sectors using their geographical region as the sensitive attribute.

\item \textbf{\nba}: This dataset includes data on approximately $400$ NBA players, using nationality as the sensitive attribute. We use this data to construct a social graph based on players' Twitter interactions. And the task is to predict whether a player's salary is above or below the median.

\end{itemize}

\noindent\textbf{Introduction on Deletion Requests.}\label{app:delete_req} To simulate real-world deletion requests, we categorize them into three types: Privileged Group Deletion, Unprivileged Group Deletion, and Uniform Deletion:
\begin{itemize}[leftmargin=0.35cm, itemindent=.0cm, itemsep=0.0cm, topsep=0.0cm]
    \item \textit{Non-Uniform Deletion}: For this deletion, we consider two cases. 1) Delete from a \textbf{Privileged Group}: It typically refers to those who historically have been more inclined to be categorized favorably in a binary classification task within machine learning. Privilege emerges from disparities in power dynamics, and it's important to note that the same groups may not universally enjoy privilege across all contexts, even within the same society~\citep{varshney2019trustworthy}. This demographic often seeks to protect their personal data, particularly due to their extensive education, thus prioritizing the privacy of their sensitive information~\citep{upton2001strategic, eurobarometer}. 2) Deletion from an \textbf{Unprivileged Group}: In contrast to the privileged group, this refers to a group less likely to receive favorable predictions. Previous studies have shown that unprivileged groups face higher privacy risks and costs for achieving fairness in machine learning models~\citep{chang2021privacy, strobel2022data}.
    
    \item \textit{\textbf{Uniform Deletion}}: This involves deletion requests from all sensitive groups and is the most common scenario. For this deletion process, we implement random sampling from the entire demographic, ensuring that each sample has an equal chance of being selected for deletion.
\end{itemize}

\definecolor{green}{rgb}{0.1,0.1,0.1}

\subsection{Deletion on Privileged Group}\label{appendix:delete_distr}

This section reports the performance of the deletion of privileged groups with different percentages, including $10\%$ and $30\%$.

\subsubsection{10\% deletion on the privileged group}
The results are shown in~\cref{tab:FGU_privileged_10}.
\begin{table}[H]
\centering
\setlength{\tabcolsep}{1.4pt}
\caption{The utility and fairness performances of {\fgu} on {\german} with sensitive attribute gender after unlearning. We assess the performance of a $10\%$ data unlearning on the privileged group.}
\label{tab:FGU_privileged_10}
\scalebox{0.88}
{
\begin{tabular}{lcccccc}
\toprule
                    & Retrain       & Fair Retrain & {\gedit}     & {\gif}        & {\gd}        & {\fgu} \\ \midrule
ACC                & $68.5\pm1.2$  & $67.3\pm1.2$ & $65.1\pm1.2$ & $66.0\pm1.2$  & $65.3\pm1.2$ & $67.1\pm1.2$ \\ 
F1                  & $79.2\pm1.6$  & $78.4\pm0.9$ & $77.2\pm0.9$ & $77.3\pm1.2$  & $76.0\pm2.4$ & $76.9\pm1.6$ \\ 
$\triangle_{DP}$    & $16.4\pm7.3$  & $2.1\pm1.3$  & $18.9\pm1.3$ & $17.4\pm10.5$ & $21.1\pm11.6$  & $2.1\pm0.6$ \\ 
$\triangle_{EO}$    & $19.2\pm9.5$  & $3.3\pm2.5$  & $21.5\pm18.4$ & $23.7\pm12.3$& $25.3\pm12.7$  & $1.3\pm0.5$ \\
\bottomrule
\end{tabular}}
\end{table}

\subsubsection{30\% deletion on the privileged group}The results are shown in~\cref{tab:FGU_privileged_30}.
\begin{table}[H]
\centering
\setlength{\tabcolsep}{1.4pt}
\caption{The utility and fairness performances of {\fgu} on {\german} with sensitive attribute gender after unlearning. We assess the performance of a $30\%$ data unlearning on the privileged group.}
\scalebox{0.88}
{
\begin{tabular}{rcccccc}
\toprule
                    & Retrain       & Fair Retrain & {\gedit}     & {\gif}        & {\gd}        & {\fgu} \\ \midrule
ACC ($\uparrow$)                & $67.3\pm1.8$  & $66.7\pm1.2$ & $64.3\pm1.2$ & $63.2\pm1.2$  & $64.1\pm1.2$ & $65.9\pm1.2$ \\ 
F1 ($\uparrow$)                 & $78.1\pm1.4$  & $77.2\pm0.9$ & $76.8\pm0.9$ & $75.9\pm1.2$  & $75.3\pm2.4$ & $76.0\pm1.6$ \\ 
$\triangle_{DP}$ ($\downarrow$)    & $19.0\pm10.3$ & $3.0\pm2.3$  & $19.6\pm11.3$ & $20.6\pm14.5$ & $24.6\pm16.6$  & $1.3\pm1.6$ \\ 
$\triangle_{EO}$  ($\downarrow$)  & $20.5\pm8.5$  & $4.5\pm2.5$  & $20.4\pm19.4$ & $26.4\pm17.3$& $28.8\pm13.7$  & $2.4\pm2.5$ \\
\bottomrule
\end{tabular}}\label{tab:FGU_privileged_30}
\end{table}

Upon analyzing the results of deletion on the privileged group from the two tables above, several observations have surfaced:
\begin{itemize}[leftmargin=0.35cm, itemindent=.0cm, itemsep=0.0cm, topsep=0.0cm]
\item  Deleting the privileged group at the same deletion ratio has a lesser impact on overall prediction performance compared to deleting the unprivileged group, as shown in \cref{tab:main_exp} in the paper.
\item  The bias introduced by deleting the privileged group at the same deletion ratio is lower than the bias resulting from deleting the unprivileged group.
\item  {\fgu} consistently demonstrates its efficacy in mitigating bias.
\end{itemize}
\subsection{Algorithm}
The {\fgu} framework is summarized in~\cref{alg:overall_workflow}.
\begin{algorithm}[t]
\SetCommentSty{small}
\LinesNumbered
\caption{\fgu}
\label{alg:overall_workflow}

\KwIn{Training graph $\graph$,
    partition algorithm $\partition$
    }
    
    \KwOut{
    Fair unlearning model $\model'_\theta$ and importance scores of the shard models $\lambda$
    }
    
    \textbf{Graph Partition and Shard Model Training:}
    \label{algline:graph_partition_start}

        Partitioning $\graph$ into $K$ shards by $\partition$ and obtain $\graph_p=\partition(\graph) = \{\graph_1, \graph_2, \cdots, \graph_k\}$;
    
    Train shard models $\model = \{\model_1, \model_2, \cdots, \model_k\}$ on $\graph_p$ and obtain the model weights ${\theta^0}=\{\theta^0_1 , \theta^0_2, \cdots,\theta^0_k\}$ and the trained shard imporance $\lambda$ \ $\rhd$  detailed \cref{method:graph_partition_training};
    \label{algline:shard_model_train_end}

    \textbf{Graph Unlearning:}\\
    Replacing $\graph_{p}$ with $\graph'_{p} = \{\graph'_1, \graph'_2, \cdots, \graph'_K$\} which removed the request nodes and/or edges.\\
    \textbf{Debiasing:}\\
    \For{$t=0,\ldots, T-1$}{

        Aggregate ${\tilde{\theta}} = 
\sum_{k=1}^K{\lambda}_k{\theta_k}$\\
        Compute the $\fairloss_{global}$ by~\cref{eq:global_dp} and $\totalloss_{global}$ by~\cref{eq:global_loss}\\
            \For{$k=0,\ldots, K-1$}{
        Calculate $\fairloss_{k}$ by~\cref{eq:local_dp} and $\totalloss_{k}$ by~\cref{eq:local_loss}.\\
        Update $\theta_k^{t+1} = \theta_k^{t}-\eta \nabla_{\theta_{k}^t} \totalloss_{k}$ \\        
        }
         \If{$t \% t_1==0$}{
         Update $\lambda^{t+1} = \lambda^{t}-\eta \nabla_{\lambda^{t}} \totalloss_{global}$
         }
        }
    \Return{$\model'_\theta$, $\lambda$.}
    
\end{algorithm}
\subsection{Using Fairness-Aware GNNs as Shard Models}
In this experiment, we compare the proposed method with two fairness-aware GNN models as shard models, FairGNN~\citep{dai2021say}, and FairSIN~\citep{yang2024fairsin}, using direct retraining on the remaining data with the requested data removed.
\begin{table}[H]
\centering
\caption{Utility and fairness performance on {\german} and {\credit}. We evaluate the methods on the original graph (FairGNN and FairSIN) and after unlearning $10\%$ of nodes and $10\%$ of edges from the original graph (FairGNN$_u$, FairAdj$_u$, and {\fgu}). The best results for the unlearning tasks are highlighted in bold.}
\setlength{\tabcolsep}{4pt}
\renewcommand{\arraystretch}{1.0}
\label{tab:fair_gnn2}
\begin{tabular}{lc|cc|ccc}
\toprule
&Method     & ACC ($\uparrow$)  & F1 ($\uparrow$)  & $\Delta_{DP}$ ($\downarrow$) & $\Delta_{EO}$ ($\downarrow$)    \\ \midrule
\multirow{5}{*}{\rotatebox{90}{{\german}}}  
& FairGNN       	& \ms{69.7}{0.3}  			&  \ms{81.6}{0.7}   			& \ms{3.5}{2.2}   & \ms{3.4}{2.2}     \\
& FairSIN 	    	 & \ms{70.1}{0.2}  			& \ms{82.5}{0.6} 			& \ms{1.6}{1.0}    & \ms{1.8}{0.8}     \\\cmidrule(r){2-6}
& FairGNN$_u$  & \ms{65.3}{1.6}  			&  \ms{75.4}{1.3}   			& \ms{4.3}{2.7}   & \ms{3.8}{3.0}     \\
& FairSIN$_u$ 	& \ms{\textbf{66.1}}{1.3}  			& \ms{\textbf{77.4}}{1.2}    	& \ms{2.4}{2.1}    & \ms{1.9}{0.9}     \\
& {\fgu}              & \ms{66.8}{1.1}  			& \ms{77.1}{2.1}   			& \ms{\textbf{2.1}}{0.5}   & \ms{\textbf{1.3}}{0.4}   \\
\midrule
\multirow{5}{*}{\rotatebox{90}{{\credit}}}  
& FairGNN       		& \ms{73.4}{1.2}  	&  \ms{81.8}{1.2}   & \ms{5.6}{2.1}   & \ms{3.1}{2.0}     \\
& FairSIN        		& \ms{77.9}{0.8}  	& \ms{87.8}{0.4}    & \ms{2.4}{1.0}  & \ms{1.7}{1.2}     \\\cmidrule(r){2-6}
& FairGNN$_u$       & \ms{66.1}{2.2}  	&  \ms{75.6}{1.6}   & \ms{6.2}{3.4}   & \ms{4.4}{3.6}     \\
& FairSIN$_u$        & \ms{\textbf{68.1}}{1.0}  		& \ms{\textbf{78.4}}{0.8}    & \ms{4.1}{2.7}  & \ms{2.6}{1.5}     \\
& {\fgu}           	    & \ms{67.5}{2.3} 		&  \ms{77.6}{1.9}   & \ms{\textbf{3.7}}{2.8} & \ms{\textbf{2.3}}{1.1}   \\
\bottomrule
\end{tabular}
\vspace{-1em}
\end{table}

From the results in~\cref{tab:fair_gnn2}, we observe that although FairSIN achieves the best performance in utility, {\fgu} achieves comparable performance. In terms of fairness, {\fgu} consistently performs the best. Without debiasing in global alignment, directly aggregating the shard models, including the ones retrained, as in FairGNN$_u$ and FairSIN$_u$, has a limited effect on debiasing. This may be because these fairness-aware algorithms struggle to converge on local data due to their complex design of the losses, thus requiring larger batch sizes, which are not suitable for the unlearning task that necessitates maintaining the size of each shard model.

\subsection{Application to the Link Prediction Task}
We consider applying our framework to another graph task, the link prediction task~\citep{kumar2020link,wang2014link,spinelli2021fairdrop,liu2024promoting}. The link prediction task involves predicting the probability that a link exists between two nodes. In the context of fairness, fair link prediction requires that the model's decision regarding the relationship between two instances is independent of their sensitive attributes~\citep{li2021dyadic}. We introduce the experimental setups in the following.

\noindent\textbf{Experimental setups.} The goal of this experiment is to unlearn a mix of nodes and links, as in the main experiment in our paper. The datasets we tested are {\cora} and {\citeseer}. The details of the datasets are in~\cref{tab:add_data}. 
\begin{table}[H]
\setlength{\tabcolsep}{3.5pt}
\centering
\caption{Statistics and properties of \cora and \citeseer.}
\begin{tabular}{lc c c c c}
\toprule
{Dataset} & {\# Nodes} &{\# Feat.} & {\# Edges} &
{Train/Valid/Test}  & {\# Classes} \\
\midrule 
\cora & $2,708$ & $1,433$ & $5,429$  & $140$/$500$/$1,000$ & $7$\\
\citeseer & $3,327$ & $3,703$ & $4,732$  & $120$/$500$/$1,000$ & $6$\\
\bottomrule
\end{tabular}
\label{tab:add_data}
\end{table}
Following a previous study~\citep{li2021dyadic}, for the fairness metrics, we use $\Delta_{DP}$, \textit{the maximum gap of true negative rate (TNR)} and \textit{the maximum gap of false negative rate (FNR)}, calculated as $\Delta_{{TNR}}:=\max _\tau\left|F_0^{\text {intra }}(\tau)-F_0^{\text {inter }}(\tau)\right|$ and $\Delta_{{FNR}}:=$ $\max _\tau\left|F_1^{\text {intra }}(\tau)-F_1^{\text {inter }}(\tau)\right|$. For the utility of link prediction, we use Area Under the Curve (AUC) and Average Precision (AP). 

As in the main experiment, the number of shards is set to $S=20$, and the batch size is set to $32$. For the baseline methods of VGAE, Fairwalk, and FairAdj, we follow the implementation details provided in~\citep{li2021dyadic}. Specifically, in FairAdj, $T_1$ is set to $50$, the total number of epochs, which includes both $T_1$ and $T_2$, is set to $4$, and a two-layer GNN with a batch size of $32$ and a learning rate of $0.01$ is applied. We use the implementation in~\footnote{https://github.com/brandeis-machine-learning/FairAdj}. For the unlearning task, we randomly sample and unlearn nodes and edges from the original data. We compare our method with one unlearning baseline, which is a variant of FairAdj, referred to as FairAdj$_u$. FairAdj$_u$ retrain on the remaining dataset after removing the requested data.

\begin{table}[H]
\centering
\caption{Utility and fairness performance on {\cora} and {\citeseer} for fair link prediction. We evaluate the methods on the original graph (VGAE, Fairwalk, and FairAdj), and after unlearning $10\%$ of nodes and $10\%$ of edges in the original graph (FairAdj$_u$ and our method, {\fgu}).}
\setlength{\tabcolsep}{2pt}
\renewcommand{\arraystretch}{1.1}
\label{tab:link_pre}
\scalebox{1.0}
{
\begin{tabular}{lc|cc|cccc}
\toprule
&Method     & AUC ($\uparrow$)  & AP ($\uparrow$)  & $\Delta_{DP}$ ($\downarrow$) & $\Delta_{{TNR}}$ ($\downarrow$) & $\Delta_{{FNR}}$ ($\downarrow$)     \\ \midrule
\multirow{5}{*}{\rotatebox{90}{{\cora}}}  
& VGAE       & \ms{\textbf{88.5}}{0.9}  	&  \ms{\textbf{90.1}}{0.8}   & \ms{27.0}{1.5}   & \ms{28.3}{5.0}     & \ms{27.1}{3.9}    \\
& Fairwalk  & \ms{88.0}{0.8}  	& \ms{88.2}{1.2}    & \ms{40.5}{2.6}  & \ms{23.8}{4.9}     & \ms{33.8}{5.1}   \\
& FairAdj   & \ms{83.9}{1.1}  	& \ms{87.1}{0.9}    & \ms{17.9}{1.2}   & \ms{16.8}{4.9}     & \ms{\textbf{15.4}}{4.0}   \\\cmidrule(r){2-7}
& FairAdj$_u$    & \ms{79.4}{1.6}  	& \ms{85.3}{1.5}    & \ms{17.1}{6.5}   & \ms{17.2}{6.1}     & \ms{17.4}{5.2}   \\
&{\fgu}        & \ms{80.3}{1.4} 	&  \ms{84.8}{1.7}   & \ms{\textbf{15.3}}{4.1}    & \ms{\textbf{16.1}}{3.7}     & \ms{16.5}{3.1}   \\
\midrule
\multirow{5}{*}{\rotatebox{90}{\citeseer}}  
& VGAE       & \ms{\textbf{81.5}}{1.2}   	& \ms{\textbf{85.6}}{1.5}   & \ms{11.4}{1.7}   & \ms{11.2}{3.7}     & \ms{11.5}{3.2}    \\
& Fairwalk  & \ms{81.3}{1.8}  	& \ms{85.1}{1.6}    & \ms{12.9}{2.8}  &  \ms{\textbf{8.1}}{3.5}     & \ms{11.4}{2.8}   \\
& FairAdj    & \ms{78.9}{1.6}  	& \ms{83.1}{2.0}    &  \ms{\textbf{8.2}}{1.5}   &  \ms{9.3}{3.1}     &  \ms{10.3}{3.0}   \\\cmidrule(r){2-7}
& FairAdj$_u$    & \ms{77.1}{1.3}  	      & \ms{80.2}{2.4}    &  \ms{10.1}{3.6}   &  \ms{9.5}{2.8}     & \ms{11.0}{2.5}   \\
&{\fgu}        & \ms{78.4}{2.1}  	& \ms{80.8}{2.0}   & \ms{8.4}{2.3}    	& \ms{8.3}{3.8}       & \ms{\textbf{9.9}}{3.1}   \\
\bottomrule
\end{tabular}}
\end{table}
From~\cref{tab:link_pre}, we observe that when unlearning $10\%$ of nodes and $10\%$ of edges, the proposed FGU achieves performance comparable to FairAdj$_u$ in terms of both utility and fairness, despite FairAdj$_u$ being trained on the original graph. This validates the effectiveness of the bi-level debiasing and aggregation design in FGU for the link prediction task.

\subsection{More baselines}
We additionally evaluated two more baselines, GUIDE~\citep{wang2023inductive} and MEGU~\citep{li2024towards}, on the {\german} dataset.

\begin{table}[H]
\centering
\caption{Additional experiments on \german. We evaluate each method after unlearning $10\%$ of nodes and $10\%$ of edges from the original graph. The best results per task are in bold.}
\setlength{\tabcolsep}{6pt}
\renewcommand{\arraystretch}{1.0}
\label{tab:fair_gnn}

\begin{tabular}{lc|cc|ccc}
\toprule
&Method     & ACC ($\uparrow$)  & F1 ($\uparrow$)  & $\Delta_{DP}$ ($\downarrow$) & $\Delta_{EO}$ ($\downarrow$)    \\ \midrule
\multirow{3}{*}{\rotatebox{90}{{\german}}}  
& GUIDE       	& \ms{65.3}{1.4}  			&  \ms{75.4}{2.2}   			& \ms{46.9}{19.4}   & \ms{29.5}{16.5}     \\
& MEGU 	    	 & \ms{66.5}{1.2}  			& \ms{\textbf{77.3}}{2.3} 			& \ms{43.6}{9.2}    & \ms{32.8}{10.4}     \\
& {FGU}              & \ms{\textbf{66.8}}{1.1}  			& \ms{77.1}{2.1}   			& \ms{\textbf{2.1}}{0.5}   & \ms{\textbf{1.3}}{0.4}  \\
\bottomrule
\end{tabular}
\end{table}

\subsection{Experimental Setup Details} \label{appendix:detail_setup}
We implement all models using the PyTorch Geometric library~\citep{paszke2017automatic} and use two-layer GNN models for our experiments. The training epochs are set to $100$ for the {\bail} dataset and $500$ for the other datasets. We use the Adam optimizer~\citep{kingma2014adam} with a default learning rate of $1\text{e-3}$ and a weight decay of $0.001$. For the hyperparameters $\alpha$ and $\beta$ of {\fgu}, we perform a search in the range $\{0.001, 0.01, 0.1, 0.5, 1, 5, 10\}$ on the validation set, ultimately selecting $\alpha=0.5$ and $\beta=1$ for the experiments. The parameters $\alpha_k$ and $\beta_k$ for the shard model are also tuned locally.

All experiments were executed on a Linux machine with 48 cores, 376GB of system memory, and two NVIDIA Tesla P100 GPUs with 12GB of GPU memory each. Details about all datasets are provided in~\cref{tab:dataset}. The data splits are public and obtained from PyTorch Geometric~\citep{fey2019fast}.

Since different methods have distinct model architectures, their hyperparameters vary and are described as follows:
\begin{itemize}[leftmargin=0.35cm, itemindent=.0cm, itemsep=0.0cm, topsep=0.0cm]
    \item \textbf{GCN}: Number of layers $\{1, 2, 3\}$, number of hidden units $\{64, 128, 256, 512\}$, learning rate $\{1\text{e-2}, 1\text{e-3}, 1\text{e-4}\}$, weight decay $\{1\text{e-4}, 1\text{e-5}\}$, dropout $\{0, 0.5, 0.8\}$.
    \item \textbf{FairGNN}: Number of hidden units 64, learning rate $\{1\text{e-2}, 1\text{e-3}, 1\text{e-4}\}$, weight decay $\{1\text{e-4}, 1\text{e-5}\}$, dropout $\{0, 0.5, 0.8\}$, hyperparameters $\alpha \{4, 5, 50, 100\}$ and $\beta \{0.01, 1, 5, 20\}$.
    \item \textbf{NIFTY}: Number of hidden units 16, project hidden units 16, drop edge rate $0.001$, drop feature rate 0.1, learning rate $\{1\text{e-2}, 1\text{e-3}, 1\text{e-4}\}$, weight decay $\{1\text{e-4}, 1\text{e-5}\}$, dropout $\{0, 0.5, 0.8\}$, regularization coefficient $\{0.2, 0.4, 0.6, 0.8\}$.
\end{itemize}

\end{document}